\documentclass[journal]{IEEEtran}
\usepackage{graphicx}
\usepackage{cite}
\usepackage{amssymb}
\usepackage{amsmath}
\usepackage{enumerate}
\usepackage{booktabs}
\usepackage{algorithm}
\usepackage{algorithmic}
\usepackage{color}
\usepackage{colortbl}
\usepackage[colorlinks,linkcolor=red,anchorcolor=blue,citecolor=green]{hyperref}
\usepackage{bbm}
\usepackage{tabularx}
\usepackage{float}
\usepackage{subfloat}
\usepackage[caption=false,font=normalsize,labelfont=sf,textfont=sf]{subfig}
\usepackage{multirow}
\usepackage[table,xcdraw]{xcolor}

\newcommand{\ie}{\textit{i.e., }}
\newcommand{\eg}{\textit{e.g., }}
\newcommand{\etal}{\textit{et al.}}

\newcolumntype{C}{>{\centering\arraybackslash}X}

\ifCLASSINFOpdf
\else
\fi

\begin{document}

%
\title{AI Security for Geoscience and Remote Sensing: Challenges and Future Trends}

\author{Yonghao~Xu,~\IEEEmembership{Member,~IEEE,}~Tao~Bai,~Weikang~Yu,~\IEEEmembership{Student~Member,~IEEE,}~Shizhen~Chang,~\IEEEmembership{Member,~IEEE,}~\\Peter~M.~Atkinson,~and~Pedram~Ghamisi,~\IEEEmembership{Senior Member,~IEEE}
\thanks{Y. Xu, S. Chang, and P. Ghamisi are with the Institute of Advanced Research in Artificial Intelligence (IARAI), 1030 Vienna, Austria (e-mail: yonghao.xu@iarai.ac.at; shizhen.chang@iarai.ac.at; pedram.ghamisi@iarai.ac.at).}
\thanks{T. Bai is with the School of Electrical and Electronic Engineering, Nanyang Technological University, Singapore 639798 (e-mail: tao.bai@ntu.edu.sg).}
\thanks{W. Yu is with Helmholtz-Zentrum Dresden-Rossendorf, Helmholtz Institute Freiberg for Resource Technology, Machine Learning Group, 09599 Freiberg, Germany (e-mail: w.yu@hzdr.de).}
\thanks{P. M. Atkinson is with the Faculty of Science and Technology, Lancaster University, LA1 4YR Lancaster, U.K., and also with the Department of Geography and Environment, University of Southampton, SO17 1BJ Southampton, U.K. (e-mail: pma@lancaster.ac.uk).}
\thanks{P. Ghamisi is also with Helmholtz-Zentrum Dresden-Rossendorf, Helmholtz Institute Freiberg for Resource Technology, Machine Learning Group, 09599 Freiberg, Germany (e-mail: p.ghamisi@hzdr.de).}}

\markboth{IEEE GEOSCIENCE AND REMOTE SENSING MAGAZINE, June~2023}%
{Shell \MakeLowercase{\textit{et al.}}: Bare Demo of IEEEtran.cls for IEEE Journals}
%

\maketitle

\begin{abstract}
Recent advances in artificial intelligence (AI) have significantly intensified research in the geoscience and remote sensing (RS) field. AI algorithms, especially deep learning-based ones, have been developed and applied widely to RS data analysis. The successful application of AI covers almost all aspects of Earth observation (EO) missions, from low-level vision tasks like super-resolution, denoising and inpainting, to high-level vision tasks like scene classification, object detection and semantic segmentation. While AI techniques enable researchers to observe and understand the Earth more accurately, the vulnerability and uncertainty of AI models deserve further attention, considering that many geoscience and RS tasks are highly safety-critical. This paper reviews the current development of AI security in the geoscience and RS field, covering the following five important aspects: adversarial attack, backdoor attack, federated learning, uncertainty and explainability. Moreover, the potential opportunities and trends are discussed to provide insights for future research. To the best of the authors' knowledge, this paper is the first attempt to provide a systematic review of AI security-related research in the geoscience and RS community. Available code and datasets are also listed in the paper to move this vibrant field of research forward.
\end{abstract}


%
\IEEEpeerreviewmaketitle

\section{Introduction}

\begin{figure*}
  \centering
  \includegraphics[width=\linewidth]{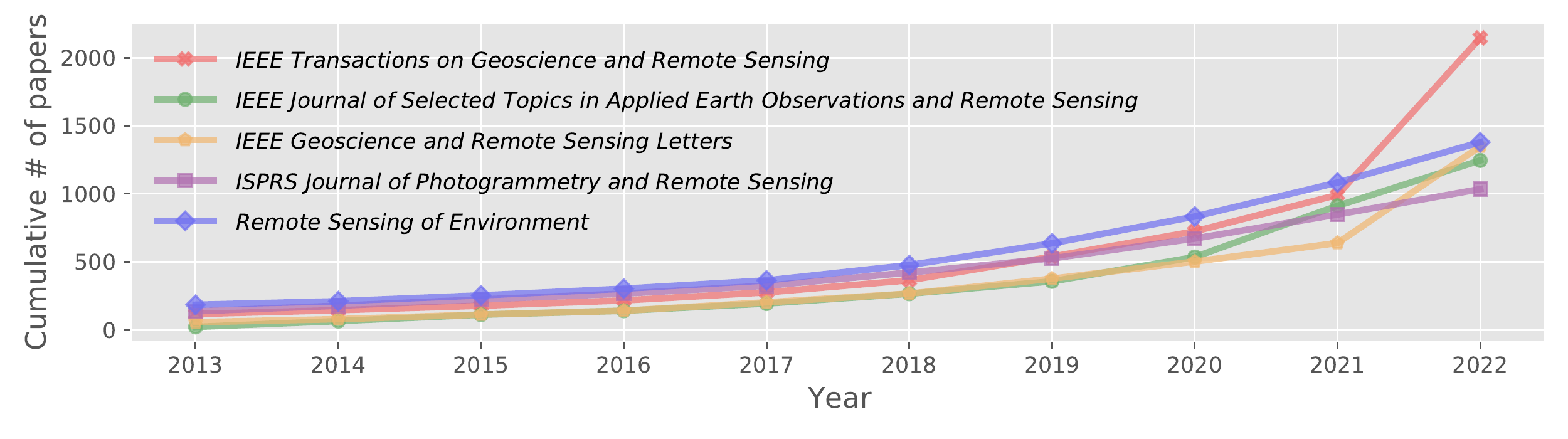}
  \caption{Cumulative numbers of artificial intelligence-related papers published in IEEE GRSS publications along with \textit{ISPRS Journal of Photogrammetry and Remote Sensing}, and \textit{Remote Sensing of Environment} in the past 10 years. The statistics are obtained from IEEE Xplore and ScienceDirect.}
\label{fig:num_paper}
\end{figure*}

\IEEEPARstart{W}{ith} the successful launch of an increasing number of remote sensing (RS) satellites, the volume of geoscience and RS data is on an explosive growth trend, bringing Earth observation (EO) missions into the big data era \cite{reichstein2019deep}. The availability of large-scale RS data has two substantial impacts: it dramatically enriches the way Earth is observed, while also demanding greater requirements for fast, accurate, and automated EO technology \cite{ghamisi2018new}. With the vigorous development and stunning achievements of artificial intelligence (AI) in the computer vision field, an increasing number of researchers are applying state-of-the-art AI techniques to numerous challenges in EO \cite{zhang2016deep}. Fig.~\ref{fig:num_paper} shows the cumulative number of AI-related papers appearing in IEEE Geoscience and Remote Sensing Society (GRSS) publications along with \textit{ISPRS Journal of Photogrammetry and Remote Sensing} and \textit{Remote Sensing of Environment} in the past ten years. It is clearly apparent that the number of AI-related papers increased significantly after 2021. The successful application of AI covers almost all aspects of EO missions, from low-level vision tasks like super-resolution, denoising and inpainting, to high-level vision tasks like scene classification, object detection and semantic segmentation \cite{zhu2017deep}. Table~\ref{tab:tasks} summarizes some of the most representative tasks in the geoscience and RS field using AI techniques and reveals the increasing importance of deep learning methods such as convolutional neural networks (CNNs) in EO.

\begin{figure}
  \centering
  \includegraphics[width=\linewidth]{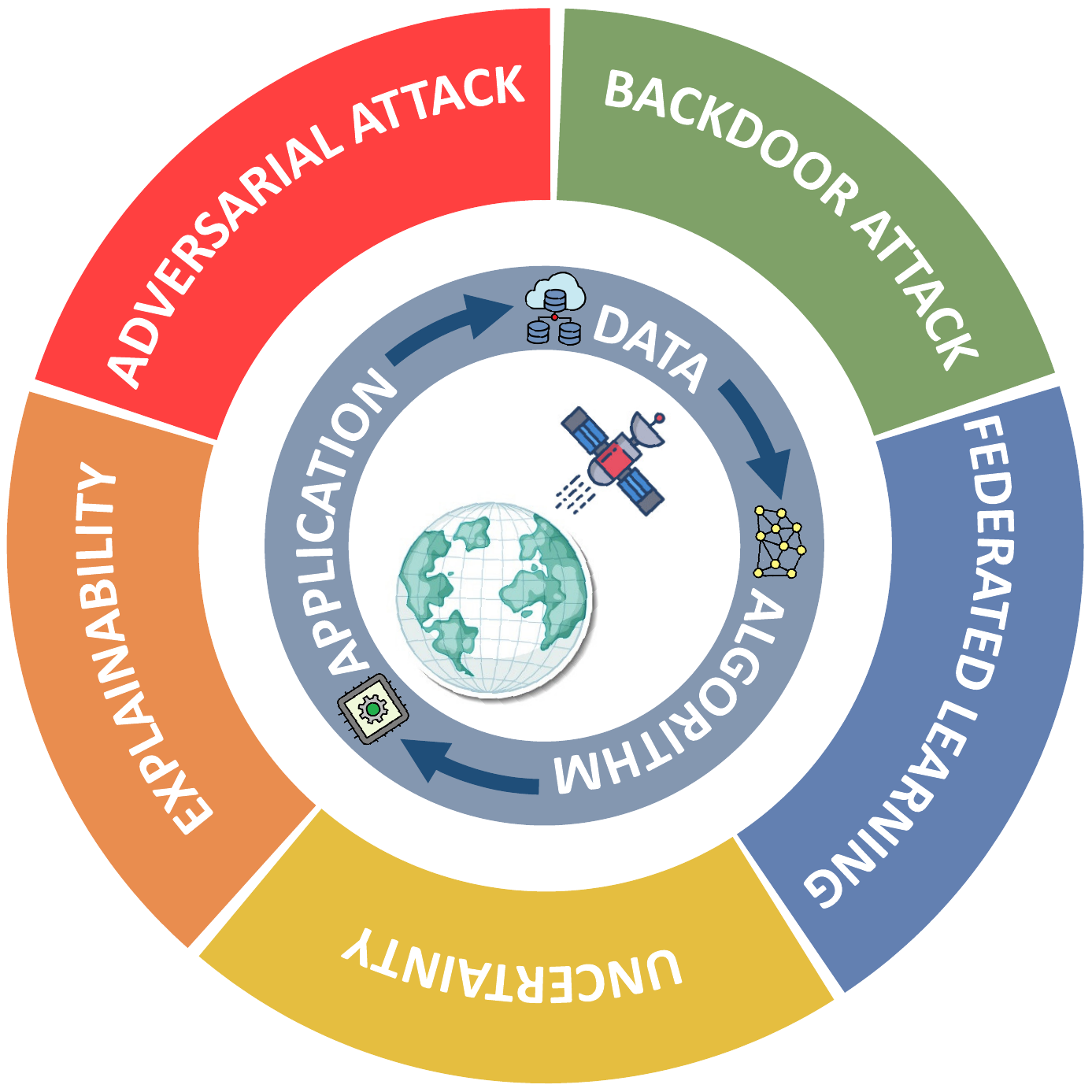}
  \caption{Overview of the research topics covered in this paper.}
\label{fig:overview}
\end{figure}

\begin{table*}
\caption{Representative Tasks in the Geoscience and Remote Sensing Field using Artificial Intelligence Techniques}
\centering
\begin{tabular}{cccc}
\toprule
\textbf{Task Types} & \textbf{AI Techniques} & \textbf{Data} & \textbf{Reference} \\
\hline
\textit{Low-Level Vision Tasks}&&&\\
Pan-sharpening&GAN&WorldView-2 and GF-2 images&\cite{ma2020pan}\\
Denoising&LRR&HYDICE and AVIRIS data&\cite{he2022}\\
Cloud removal&CNN&Sentinel-1 and Sentinel-2 data&\cite{ebel2022sen12ms}\\
Destriping&CNN&EO-1 Hyperion and HJ-1A images&\cite{zhong2020satellite}\\
\hline
\textit{High-Level Vision Tasks}&&&\\
Scene classification&CNN&Google Earth images&\cite{cheng2017remote}\\
Object detection&CNN&Google Earth images, GF-2, and JL-1 images&\cite{ding2021object}\\
Land use and land cover mapping&FCN&Airborne hyperspectral/VHR color image/LiDAR data&\cite{xu2019advanced}\\
Change detection&SN and RNN&GF-2 images&\cite{chen2019change}\\
Video tracking&SN and GMM&VHR satellite videos&\cite{shao2021hrsiam}\\
\hline
\textit{Natural Language Processing-Related Tasks}&&&\\
Image captioning&RNN&VHR satellite images with text descriptions&\cite{lu2017exploring}\\
Text-to-image generation&MHN&VHR satellite images with text descriptions&\cite{xu2022txt2img}\\
Visual question answering&CNN and RNN&Satellite/aerial images with visual questions/answers&\cite{lobry2020rsvqa}\\
\hline
\textit{Environment Monitoring Tasks}&&&\\
Wildfire detection&FCN&Sentinel-1, Sentinel-2, Sentinel-3 and MODIS data&\cite{rashkovetsky2021wildfire}\\
Landslide detection&FCN and Transformer&Sentinel-2 and ALOS PALSAR data&\cite{ghorbanzadeh2022outcome}\\
Weather forecasting&CNN and LSTM&SEVIRI data&\cite{dewitte2021artificial}\\
Air quality prediction&ANN&MODIS data&\cite{feng2019neural}\\
Poverty estimation&CNN&VHR satellite images&\cite{jean2016combining}\\
Refugee camps detection&CNN&WorldView-2 and WorldView-3 data&\cite{gella2022mapping}\\
\bottomrule
\end{tabular}
\label{tab:tasks}
\end{table*}

Despite the great success achieved by AI techniques, related safety and security issues should not be neglected \cite{zhang2022artificial}. While advanced AI models like CNNs possess powerful data fitting capabilities and are designed to learn like the human brain, they usually act like black boxes, which makes it difficult to understand and explain how they work \cite{ge2022geoscience}. Moreover, such characteristics may lead to uncertainties, vulnerabilities and security risks, which could seriously threaten the safety and robustness of the geoscience and RS tasks. Considering that most of these tasks are highly safety-critical, this paper aims to provide a systematic review of current developments in AI security in the geoscience and RS field. As shown in Fig.~\ref{fig:overview}, the main research topics covered in this paper comprise the following five important aspects: adversarial attack, backdoor attack, federated learning, uncertainty and explainability. A brief introduction for each topic is given below:
\begin{itemize}
    \item \textit{Adversarial attack} focuses on attacking the inference stage of a machine learning model by generating adversarial examples. Such adversarial examples may look identical to the original clean samples but can mislead the machine learning models to yield incorrect predictions with high confidence.
    \item \textit{Backdoor attack} aims to conduct data poisoning with specific triggers in the training stage of a machine learning model. The infected model may yield normal predictions on benign samples but make specific incorrect predictions on samples with backdoor triggers.
    \item \textit{Federated learning} ensures data privacy and data security by training machine learning models with decentralized data samples without sharing data.
    \item \textit{Uncertainty} aims to estimate the confidence and robustness of the decisions made by machine learning models.
    \item \textit{Explainability} aims to provide an understanding of and interpret, machine learning models, especially black-box models like CNNs.
\end{itemize}

While research on the above topics is still in its infancy in the geoscience and RS field, these topics are indispensable for building a secure and trustworthy EO system.

\begin{table}
\caption{Main Abbreviations and Nomenclatures}
\label{tab:abbrev}
\centering
\scriptsize
\begin{tabular}{cl}
\toprule
\textbf{Abbrev./Notation}  &  \textbf{Definition}\\
\midrule
Adam & Adaptive moment estimation\\
AdamW & Adam with decoupled weight decay\\
AI  & Artificial intelligence \\
ANN & Artificial neural network \\
BNN & Bayesian neural network \\
CE & Cross-entropy \\
CNN & Convolutional neural network \\
DNN & Deep neural network \\
EO  & Earth observation \\
FCN & Fully convolutional network \\
FL & Federated learning \\
GAN & Generative adversarial network \\
GMM & Gaussian mixture model \\
HSI & Hyperspectral image \\
IoT & Internet of things \\
LRR & Low-rank representation \\
LSTM & Long short-term memory network \\
MHN & Modern Hopfield network \\
ML & Machine learning \\
PM$_{2.5}$ & Particulate matter with a diameter of 2.5 $\mu$m or less \\
RGB & Red-green-blue \\
RNN & Recurrent neural network \\
RS  & Remote sensing \\
SAR & Synthetic aperture radar \\
SGD & Stochastic gradient descent \\
SN & Siamese network \\
SVM & Support vector machine \\
UAV & Unmanned aerial vehicle \\
VHR & Very high-resolution \\
XAI & Explainable AI \\
\midrule
$f$  & The classifier mapping of a neural network model \\
$\boldsymbol{\theta}$  & A set of parameters in a neural network model \\
$\mathcal{X}$  & The image space \\
$\mathcal{Y}$  & The label space \\
$x$  & A sample from the image space \\
$y$  & The corresponding label of $x$ \\
$\hat{y}$  & The predicted label of $x$ \\
$\delta$ & The perturbation in adversarial examples \\
$\epsilon$ & The perturbation level of adversarial attacks \\
$\nabla$ & The gradient of the function \\
$\mathcal{D}_b$ & The benign training set \\
$\mathcal{D}_p$ & The poisoned set \\
$\mathcal{R}_b$ & The standard risk \\
$\mathcal{R}_a$ & The backdoor attack risk \\
$\mathcal{R}_p$ & The perceivable risk \\
$\boldsymbol{t}$ & The trigger patterns for backdoor attacks \\
$s$ & The sample proportion \\
$E$ & The explanation of a neural network model \\
$\mathcal{L}$ & The loss function of a neural network model \\
$\textrm{sign}(\cdot)$ & Signum function \\
$\mathbb{I}(\cdot)$ & Indicator function \\
\bottomrule
\end{tabular}
\end{table}

The main contributions of this paper are summarized as follows:
\begin{itemize}
    \item For the first time, we provide a systematic and comprehensive review of AI security-related research for the geoscience and RS community, covering five aspects: adversarial attack, backdoor attack, federated learning, uncertainty and explainability.
    \item In each aspect, a theoretical introduction is provided and several representative works are organized and described in detail, emphasizing in each case the potential connection with AI security for EO. In addition, we provide a perspective on the future outlook of each topic, to further highlight the remaining challenges in the field of geoscience and RS.
    \item We summarize the entire review with four possible research directions in EO: secure AI models, data privacy, trustworthy AI models and explainable AI models. In addition, potential opportunities and research trends are identified for each direction to arouse readers' research interest in AI security.
\end{itemize}

Table~\ref{tab:abbrev} provides the main abbreviations and nomenclatures used in this paper. The rest of this paper is organized as follows. Section~\ref{sec:adv} reviews adversarial attacks and defenses for RS data. Section~\ref{sec:back} further reviews backdoor attacks and defenses in the geoscience and RS field. Section~\ref{sec:fed} introduces the concepts and applications of federated learning in the geoscience and RS field. Section~\ref{sec:un} describes the sources of uncertainty in EO and summarizes the most commonly used methods for uncertainty quantification. Section~\ref{sec:ex} introduces representative explainable AI applications in the geoscience and RS field. Conclusions and other discussions are summarized in Section~\ref{sec:con}.

\section{Adversarial Attack}
\label{sec:adv}

\begin{figure}[t]
    \centering
    \includegraphics[width=\linewidth]{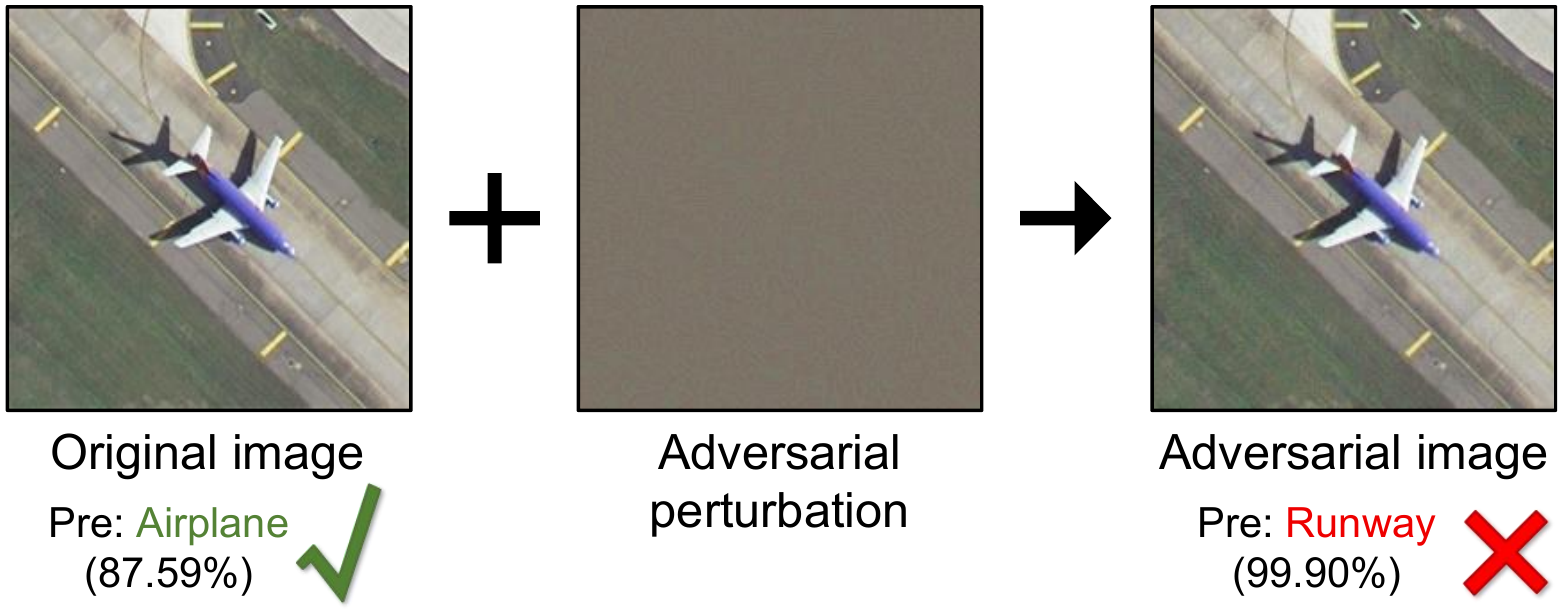}
    \caption{Adversarial attacks causing AlexNet~\cite{krizhevsky2017imagenet} to predict the VHR image from ``Airplane'' to ``Runway'' with high confidence.}
    \label{fig:adv_VHR}
\end{figure}

AI techniques have been deployed widely in geoscience and RS, as shown in Table~\ref{tab:tasks}, and achieved great success over the past decades.
The existence of adversarial examples, however, threatens such machine learning models and raises concerns about the security of these models.
With slight and imperceptible perturbations, the clean RS images can be manipulated to be adversarial and fool well-trained machine learning models, \ie making incorrect predictions~\cite{czaja2018adversarial} (see Fig.~\ref{fig:adv_VHR} for an example). 
Undoubtedly, such vulnerabilities of machine learning models are harmful and would hinder their potential for safety-critical geoscience and RS applications.
To this end, it is critical for researchers to study the vulnerabilities (\textit{adversarial attacks}) and develop corresponding methods (\textit{adversarial defenses}) to harden machine learning models for EO missions.

\subsection{Preliminaries}
Adversarial attacks usually refer to finding adversarial examples for well-trained models (target models).
Taking image classification as an example, let $f: \mathcal{X} \rightarrow \mathcal{Y}$ be a classifier mapping from the image space $\mathcal{X} \subset \mathbb{R}^d$ to the label space $\mathcal{Y}=\{ 1, \cdots, K\}$ with parameters $\boldsymbol{\theta}$, where $d$ and $K$ denote the numbers of pixels and categories, respectively.
Given the perturbation budget $\epsilon$ under $\ell_{p}$-norm, the common way to craft adversarial examples for the adversary is to find a perturbation $\delta \in \mathbb{R}^{d}$, which can maximize the loss function, \eg cross-entropy loss $\mathcal{L}_{ce}$, so that $f(x+\delta)\neq y$, where $y$ is the label of $x$. 
Therefore, $\delta$ can be obtained by solving the optimization problem as below:
\begin{equation}\begin{array}{c}
\delta^*=\mathop{\arg\max}\limits_{B(x,\epsilon)} \mathcal{L}_{ce}(\boldsymbol{\theta}, x+\delta, y),
\end{array}
\label{eqn:adv_delta}
\end{equation}
where $B(x,\epsilon)$ is the allowed perturbation set, expressed as
$B(x,\epsilon):=\left\{x+\delta \in \mathbb{R}^{d}  \mid  \|\delta\| _{p} \leq \epsilon\right\}$. 
Common values of $p$ are 0, 1, 2 and $\infty$.
In most cases, $\epsilon$ is set to be small so that the perturbations are imperceptible to human eyes.

To solve Eq.~\eqref{eqn:adv_delta}, gradient-based methods~\cite{Goodfellow_Shlens_Szegedy_2015,Kurakin_Goodfellow_Bengio_2017,Madry_Makelov_Schmidt_Tsipras_Vladu_2018} are usually exploited.
One of the most popular solutions is Projected Gradient Descent (PGD)~\cite{Madry_Makelov_Schmidt_Tsipras_Vladu_2018}, which is an iterative method.
Formally, the perturbation $\delta$ is updated in each iteration as follows:
\begin{equation}
\begin{array}{cc}
\delta^{i+1} =\operatorname{Proj}_{B(x,\epsilon)}\left(\delta^{i}+\alpha \operatorname{sign}\left(\nabla_{x^{i}} \mathcal{L}_{ce}\left(\boldsymbol{\theta}, x^{i}, y\right)\right)\right), \\
x^{i} = x + \delta^i, \delta^{0} = 0
\end{array}
\label{eqn:pgd}
\end{equation}
where $i$ is the current step, $\alpha$ is the step size (usually smaller than $\epsilon$) and $\operatorname{Proj}$ is the operation to make sure the values of $\delta$ are valid.
Specifically, for $(i+1)th$ iteration, we first calculate the gradients of $\mathcal{L}_{ce}$ \textit{w.r.t} $x^{i} = x + \delta^i$, then add the gradients to previous perturbations $\delta^i$ and obtain $\delta^{i+1}$. To further ensure that the pixel values in the generated adversarial examples are valid (\eg within $\left[0,1\right]$), the $\operatorname{Proj}$ operation is adopted to clip the intensity values in $\delta^{i+1}$.

There are different types of adversarial attacks, depending on the adversary's knowledge and goals. 
If the adversary can access the target models, including the structures, parameters, and training data, it is categorized as a \textit{white-box attack}; 
otherwise, if the adversary can access only the outputs of target models, it is known as a \textit{black-box attack}. 
When launching attacks, if the goal of the adversary is simply to fool target models so that \mbox{$f(x+\delta) \neq y$}, this is a \textit{non-targeted attack};
otherwise, the adversary expects target models to output specific results so that \mbox{$f(x+\delta) = y_t$} ($y_t\neq y$ is the label of the target class specified by the adversary), which is a \textit{targeted attack}. 
In addition, if the adversarial attacks are independent of data, such attacks are called \textit{universal attacks}.

The attack success rate is a widely adopted metric for evaluating adversarial attacks. It measures the proportion of adversarial examples that successfully deceive the target model, resulting in incorrect predictions.

\subsection{Adversarial Attacks}
Adversarial examples for deep learning were discovered initially in \cite{szegedyIntriguingPropertiesNeural2014}.
Many pioneer works on adversarial examples have appeared since that time~\cite{Goodfellow_Shlens_Szegedy_2015,Kurakin_Goodfellow_Bengio_2017,Madry_Makelov_Schmidt_Tsipras_Vladu_2018} and have motivated research on adversarial attacks on deep neural networks (DNNs) in the context of RS.
Czaja~\etal~\cite{czaja2018adversarial} revealed the existence of adversarial examples for RS data for the first time, and focused on targeted adversarial attacks for deep learning models.
They also pointed out two key challenges of designing physical adversarial examples in RS settings: viewpoint geometry and temporal variability.
Chen~\etal~\cite{chen2019adversarial} confirmed the conclusions in~\cite{czaja2018adversarial} with extensive experiments across various CNN models and adversarial attacks for RS.
Xu~\etal~\cite{xu2020assessing} further extended evaluation of the vulnerability of deep learning models to untargeted attacks, which is complementary to \cite{czaja2018adversarial}.
It was reported that most state-of-the-art DNNs can be fooled by adversarial examples with very high confidence.
The transferability of adversarial examples was first discussed in~\cite{xu2020assessing}, which indicated the harmfulness of adversarial examples generated with one specific model on other different models.
According to their experiments, adversarial examples generated by AlexNet~\cite{krizhevsky2017imagenet} can cause performance drops in deeper models of different degrees, and deep models are more resistant to adversarial attacks than shallow models.
Xu and Ghamisi~\cite{xu2022universal} exploited such transferability and developed the Universal Adversarial Examples (UAE)\footnote{\url{https://drive.google.com/file/d/1tbRSDJwhpk-uMYk2t-RUgC07x2wyUxAL/view?usp=sharing}}\footnote{\url{https://github.com/YonghaoXu/UAE-RS}}.
UAE enables adversaries to launch adversarial attacks without accessing the target models. 
Another interesting observation in~\cite{xu2020assessing} is that traditional classifiers like SVMs are less vulnerable to adversarial examples generated by DNNs.
However, this does not mean traditional classifiers are robust to adversarial examples~\cite{zhou2012adversarial}.
While the UAE is designed only for fooling target models, Bai~\etal~\cite{bai_targeted_2022} extended the UAE and developed two targeted universal attacks for specific adversarial purposes\footnote{\url{https://github.com/tao-bai/TUAE-RS}}.
It is worth noting that such targeted universal attacks sacrifice the transferability between different models, and enhancing the transferability of targeted attacks is still an open problem. 

\begin{figure}[t]
    \centering
    \includegraphics[width=\linewidth]{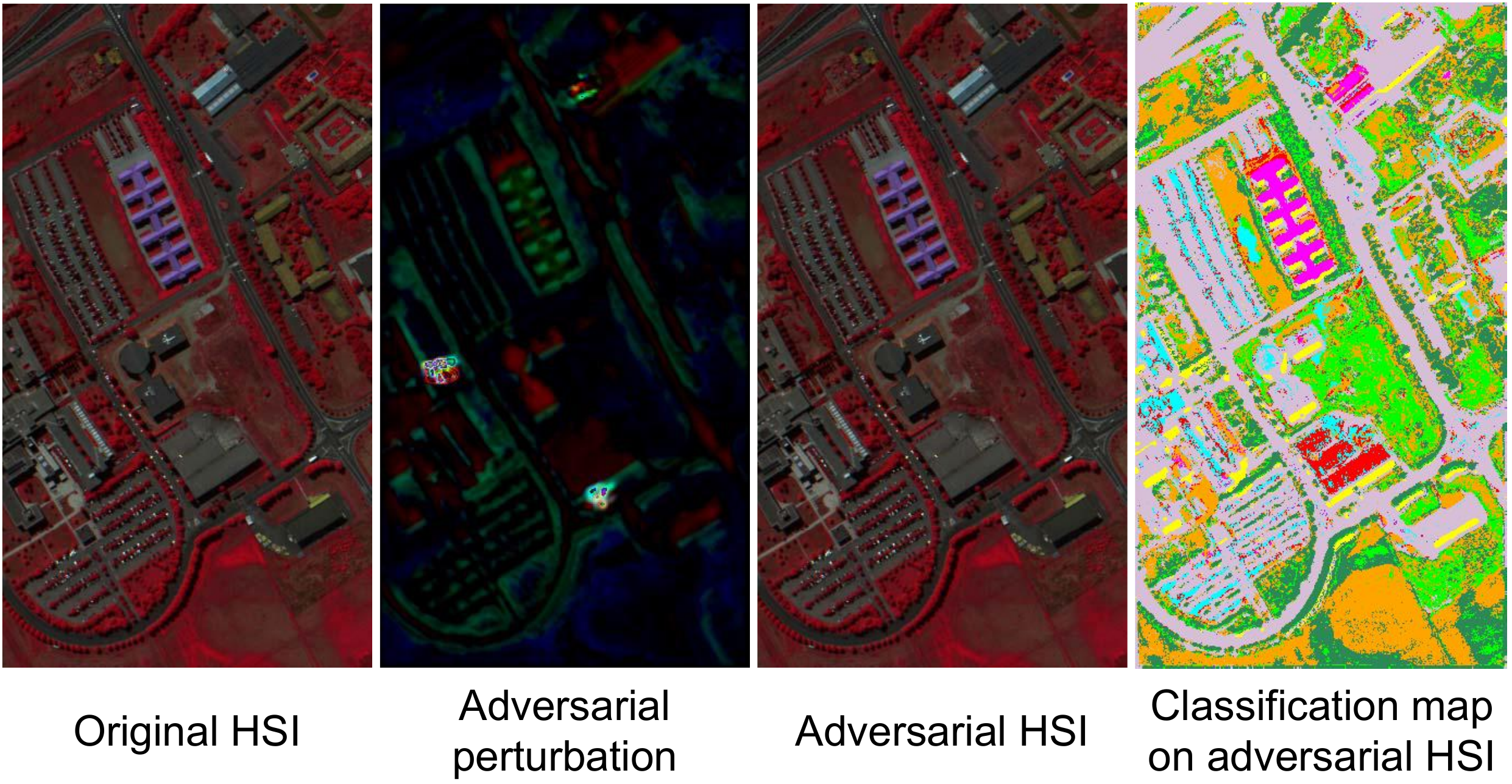}
    \caption{The threat of adversarial attacks in the hyperspectral domain \cite{xu2021self}. From left to right: original HSI (in false color), adversarial perturbation with $\epsilon= 0.04$, adversarial HSI, and the classification map on the adversarial HSI using PResNet \cite{paoletti2018deep}, which is seriously fooled (with an overall accuracy of $35.01\%$).}
    \label{fig:hsiadv}
\end{figure}

In addition to deep learning models for optical images, those for hyperspectral images~(HSIs) and synthetic aperture radar~(SAR) images in RS are also important.
For HSI classification, the threats of adversarial attacks are more serious due to the limited training data and high dimensionality~\cite{xu2021self}.
Xu~\etal~\cite{xu2021self} first revealed the existence of adversarial examples in the hyperspectral domain\footnote{\url{https://github.com/YonghaoXu/SACNet}}, which can easily compromise several state-of-the-art deep learning models (see Fig.~\ref{fig:hsiadv} for an example).
Considering the high dimensionality of HSI, ~\cite{shiHyperspectralImageClassification2022} investigated generating adversarial samples close to the decision boundaries with minimal disturbances.
Unlike optical images and HSIs with multiple dimensions, SAR images are acquired in microwave wavelengths, and contain only the backscatter information in a limited number of bands.
Chen~\etal~\cite{chen2021empirical} and Li~\etal~\cite{li_adversarial_2021} empirically investigated adversarial examples on SAR images using existing attack methods and found that the predicted classes of adversarial SAR images were highly concentrated. 
Another interesting phenomenon observed in~\cite{chen2021empirical} is that adversarial examples generated on SAR images tend to have greater transferability between different models than optical images, which indicates that SAR recognition models are easier to attack and raises security concerns when applying SAR data in EO missions.




\subsection{Adversarial Defenses}
Adversarial attacks reveal the drawbacks of current deep learning-based systems for EO and raise public concerns about RS applications.
Thus, it is urgent to develop corresponding adversarial defenses against such attacks and avoid severe consequences.
Adversarial training~\cite{bai_recent_2021} is recognized as one of the most effective adversarial defenses against adversarial examples and has been applied widely in computer vision tasks.
The idea behind adversarial training is intuitive: it trains directly deep learning models on adversarial examples generated in each loop.
Xu~\etal~\cite{xu2020assessing} took the first step and empirically demonstrated adversarial training for the RS scene classification task.
Their extensive experiments showed that adversarial training increased significantly the resistance of deep models to adversarial examples, although evaluation was limited to the naive attack Fast Gradient Descent Method~(FGSM)~\cite{Goodfellow_Shlens_Szegedy_2015}. 
Similar methods and conclusions were also obtained in~\cite{chan2021demotivate,peng2022scattering}.
However, adversarial training requires labeled data and suffers significant decreases in accuracy on testing data~\cite{xu2021robust}.
Xu~\etal~\cite{xu2021robust} introduced self-supervised learning into adversarial training to extend the training set with unlabeled data to train more robust models.
Cheng~\etal~\cite{9442932} proposed another variant of adversarial training, where a generator is utilized to model the distributions of adversarial perturbations. 
Unlike the aforementioned research, which used mainly the adversarial training technique, some further attempts were made to improve adversarial robustness by modifying model architectures.
Xu~\etal~\cite{xu2021self} introduced a self-attention context network, which extracts both local and global context information simultaneously.
By extracting global context information, pixels are connected to other pixels in the whole image and obtain resistance to local perturbations.
It is also reasonable to add preprocessing modules before the original models. For example, Xu~\etal~\cite{xu2022task} proposed to purify adversarial examples using a denoising network.
Since the adversarial examples and original images have different distributions, such discrepancies have inspired researchers to develop methods to detect adversarial examples. 
Chen~\etal~\cite{9492037} noticed the class selectivity of adversarial examples, (\ie the misclassified classes are not random).
They compared the confidence scores of original samples and adversarial examples and obtained class-wise soft thresholds for use as an indicator for adversarial detection.
Similarly, from the energy perspective, Zhang~\etal~\cite{zhang2022energy} captured an inherent energy gap between the adversarial examples and original samples.

\subsection{Future Perspectives}
Although much research related to security issues in RS was discussed above, the threats from adversarial examples have not been eliminated completely.
Here, we summarize some potential directions for studying adversarial examples:
\subsubsection{Adversarial attacks and defenses beyond scene classification}
As in the literature review introduced above, the focus of most adversarial attacks in RS is scene classification.
Many other tasks like object detection~\cite{ding2021object} and video tracking~\cite{shao2021hrsiam} remain untouched, where DNNs are deployed as widely as in scene classification.
Thus, it is equally important to study these tasks from an adversarial perspective.

\subsubsection{Different forms of adversarial attacks}
When talking about adversarial examples, we usually refer to adversarial perturbations.
Nevertheless, crafting adversarial examples is not limited to adding perturbations since the existence of adversarial examples is actually caused by the gap between human vision and machine vision.
Such gaps have not been well-defined yet, which may enable us to explore adversarial examples in different forms. 
For example, scholars have explored the use of adversarial patches~\cite{zhang2022adversarial,sun2023threatening}, where a patch is added to an input image to deceive the machine, as well as the concept of natural adversarial examples~\cite{burnel2021generating}, where an image looks the same to humans but is misclassified by the machine due to subtle differences. These approaches may offer insights into the mechanisms underlying adversarial examples. 
By better understanding the existence of adversarial examples in different forms, we can develop more comprehensive and effective defenses to protect against these attacks.

\subsubsection{Different scenarios of adversarial attacks}
Despite white-box settings being the most common setting when we discuss the robustness of DNNs, black-box settings are more practical for real-world applications, where the adversary has no or limited access to the trained models in deployment.
Typically, there are two strategies that adversaries can employ in a black-box scenario. 
The first is adversarial transferability~\cite{Dong_2018_CVPR,xie2019improving}, which involves creating a substitute model that imitates the behavior of the target model based on a limited set of queries or inputs. Once the substitute model is created, the adversary can generate adversarial examples on the substitute model and transfer these examples to the target model. 
The second strategy is to directly query the target model using input-output pairs and use the responses to generate adversarial examples. This approach is known as the query-based attack~\cite{sun2022query,sun2022exploring}. 
Future research in this area will likely focus on the development of more effective black-box attacks.

\subsubsection{Physical adversarial examples}
The current research on adversarial examples in the literature focuses on digital space without considering the physical constraints that may exist in the real world. Thus, one natural question that arises in the context of physical adversarial examples is whether the adversarial perturbations will be detectable or distorted when applied in the real world, where the imaging environment is more complex and unpredictable, leading to a reduction in their effectiveness.
Therefore, it is crucial to explore whether adversarial examples can be designed physically for specific ground objects~\cite{deng2023rust,sun2023threatening}, especially considering that many DNN-based systems are currently deployed for Earth observation (EO) missions. Incorporating physical constraints in adversarial examples may further increase our understanding of the limits of adversarial attacks in remote sensing applications and aid in developing more robust and secure systems.

\begin{table*}
\centering
\caption{Differences and Connections Between Different Types of Attacks for Machine Learning Models}
\label{tab:diff_att}
\scriptsize
\begin{tabularx}{\linewidth}{lXlXl}
\toprule
\textbf{Attack Type} & \textbf{Attack Goal} & \textbf{Attack Stage} & \textbf{Attack Pattern} & \textbf{Transferability} \\
\specialrule{0.05em}{1pt}{1pt}
Adversarial attack \cite{szegedyIntriguingPropertiesNeural2014} & Cheat the model to yield wrong predictions with specific perturbations & Evaluation phase & Various patterns calculated for different samples & Transferable \\
Data poisoning \cite{xiao2015feature} & Damage model performance with out-of-distribution data & Training phase & Various patterns selected by the attacker & Non-transferable \\
Backdoor attack \cite{gu2017badnets} & Mislead the model to yield wrong predictions on data with embedded triggers & Training phase & Fixed patterns selected by the attacker & Non-transferable
\\
\bottomrule
\end{tabularx}%
\end{table*}

\subsubsection{The positive side of adversarial examples}
Although often judged to be harmful, adversarial examples indeed reveal some intrinsic characteristics of DNNs, which more or less help us understand DNNs more deeply.
Thus, researchers should not only focus on generating solid adversarial attacks but also investigate the potential usage of adversarial examples for EO missions in the future.

\section{Backdoor Attack}
\label{sec:back}

Although adversarial attacks bring substantial security risks to machine learning models in geoscience and RS, these algorithms usually assume that the adversary can only attack the target model in the evaluation phase. In fact, applying machine learning models to RS tasks often involves multiple steps, from data collection, model selection and model training to model deployment. Each of these steps offers potential opportunities for the adversary to conduct attacks. Since acquiring high-quality annotated RS data is very time-consuming and labor-intensive, researchers may use third-party datasets directly to train machine learning models, or even use directly the pre-trained machine learning models from a third party in a real-world application scenario. In these cases, the probability of the target model being attacked during the training phase is greatly increased. One of the most representative attacks designed for the training phase is the backdoor attack, also known as the Trojan attack \cite{liu2017neural}. Table~\ref{tab:diff_att} summarizes the main differences and connections between the backdoor attack and other types of attacks for machine learning models. To help readers better understand the background of backdoor attacks, this section will first summarize the related preliminaries. Then, some representative works about backdoor attacks and defenses will be introduced. Finally, perspectives on the future of this research direction will be discussed.

\subsection{Preliminaries} 
The main goal of backdoor attacks is to induce the deep learning model to learn the mapping between the hidden backdoor triggers and the malicious target labels (specified by the attacker) by poisoning a small portion of the training data. Formally, let $f: \mathcal{X} \rightarrow \mathcal{Y}$ be a classifier mapping from the image space $\mathcal{X} \subset \mathbb{R}^d$ to the label space $\mathcal{Y}=\{ 1, \cdots, K\}$ with parameters $\boldsymbol{\theta}$, where $d$ and $K$ denote the numbers of pixels and categories, respectively. We use $\mathcal{D}_b=\{\left(\boldsymbol{x}_i,y_i\right)\}_{i=1}^N$ to represent the original benign training set, where $\boldsymbol{x}_i\in\mathcal{X}$, $y_i\in\mathcal{Y}$, and $N$ denotes the total number of sample pairs. The standard risk $R_b$ of the classifier $f$ on the benign training set $\mathcal{D}_b$ can then be defined as
\begin{equation}
R_b\left(\mathcal{D}_b\right)=\mathbb{E}_{\left(\boldsymbol{x},y\right)\sim \mathcal{P_D}}\mathbb{I}\left(\arg\max\left(f\left(\boldsymbol{x}\right)\right)\ne y\right),
    \label{riskb}
\end{equation}
where $\mathcal{P_D}$ denotes the distribution behind the benign training set $\mathcal{D}_b$, $\mathbb{I}\left(\cdot\right)$ is the indicator function (\ie $\mathbb{I}\left(\texttt{condition}\right)=1$ if and only if \texttt{condition} is true), and $\arg\max\left(f\left(\boldsymbol{x}\right)\right)$ denotes the predicted label by the classifier $f$ on the input sample $\boldsymbol{x}$. With Eq. \eqref{riskb}, we can measure whether the classifier $f$ can correctly classify the benign samples.

Let $\mathcal{D}_p$ denote the poisoned set, which is a subset of $\mathcal{D}_b$. The backdoor attack risk $R_a$ of the classifier $f$ on the poisoned set $\mathcal{D}_p$ can then be defined as
\begin{equation}
R_a\left(\mathcal{D}_p\right)=\mathbb{E}_{\left(\boldsymbol{x},y\right)\sim \mathcal{P_D}}\mathbb{I}\left(\arg\max\left(f\left(G_{\boldsymbol{t}}\left(\boldsymbol{x}\right)\right)\right)\ne S\left(y\right)\right),
 \label{riska}
\end{equation}
where $G_{\boldsymbol{t}}\left(\cdot\right)$ denotes an injection function that injects the trigger patterns $\boldsymbol{t}$ specified by the attack to the input benign image, and $S\left(\cdot\right)$ denotes the label shifting function that maps the original label to a specific category specified by the attack. With Eq. \eqref{riska}, we can measure whether the attacker can successfully trigger the classifier $f$ to yield malicious predictions on the poisoned samples.

Since backdoor attacks aim to achieve imperceptible data poisoning, the perceivable risk $R_p$ is further defined as
\begin{equation}
R_p\left(\mathcal{D}_p\right)=\mathbb{E}_{\left(\boldsymbol{x},y\right)\sim \mathcal{P_D}}\mathbb{I}\left(C\left(G_{\boldsymbol{t}}\left(\boldsymbol{x}\right)\right)=1\right),
    \label{riskp}
\end{equation}
where $C\left(\cdot\right)$ denotes a detector function, and $C\left(G_{\boldsymbol{t}}\left(\boldsymbol{x}\right)\right)=1$ if and only if the poisoned sample $G_{\boldsymbol{t}}\left(\boldsymbol{x}\right)$ can be detected as an abnormal sample. With Eq. \eqref{riskp}, we can measure how stealthy the backdoor attacks could be.

Based on the aforementioned risks, the overall objective of the backdoor attacks can be summarized as
\begin{equation}
    \min_{\boldsymbol{\theta,t}} R_b\left(\mathcal{D}_b-\mathcal{D}_p\right)+\lambda_a R_a\left(\mathcal{D}_p\right)+\lambda_p R_p\left(\mathcal{D}_p\right),
    \label{risk}
\end{equation}
where $\lambda_a$ and $\lambda_p$ are two weighting parameters. Commonly, the ratio between the number of poisoned samples $|\mathcal{D}_p|$ and the number of benign samples $|\mathcal{D}_b|$ used in the training phase is called the poisoning rate $|\mathcal{D}_p|/|\mathcal{D}_b|$ \cite{li2022backdoor}.

There are two primary metrics used to evaluate backdoor attacks: attack success rate and benign accuracy. The attack success rate measures the proportion of misclassified samples on the poisoned test set, while benign accuracy measures the proportion of correctly classified benign samples on the original clean test set.

\begin{figure}
  \centering
  \includegraphics[width=\linewidth]{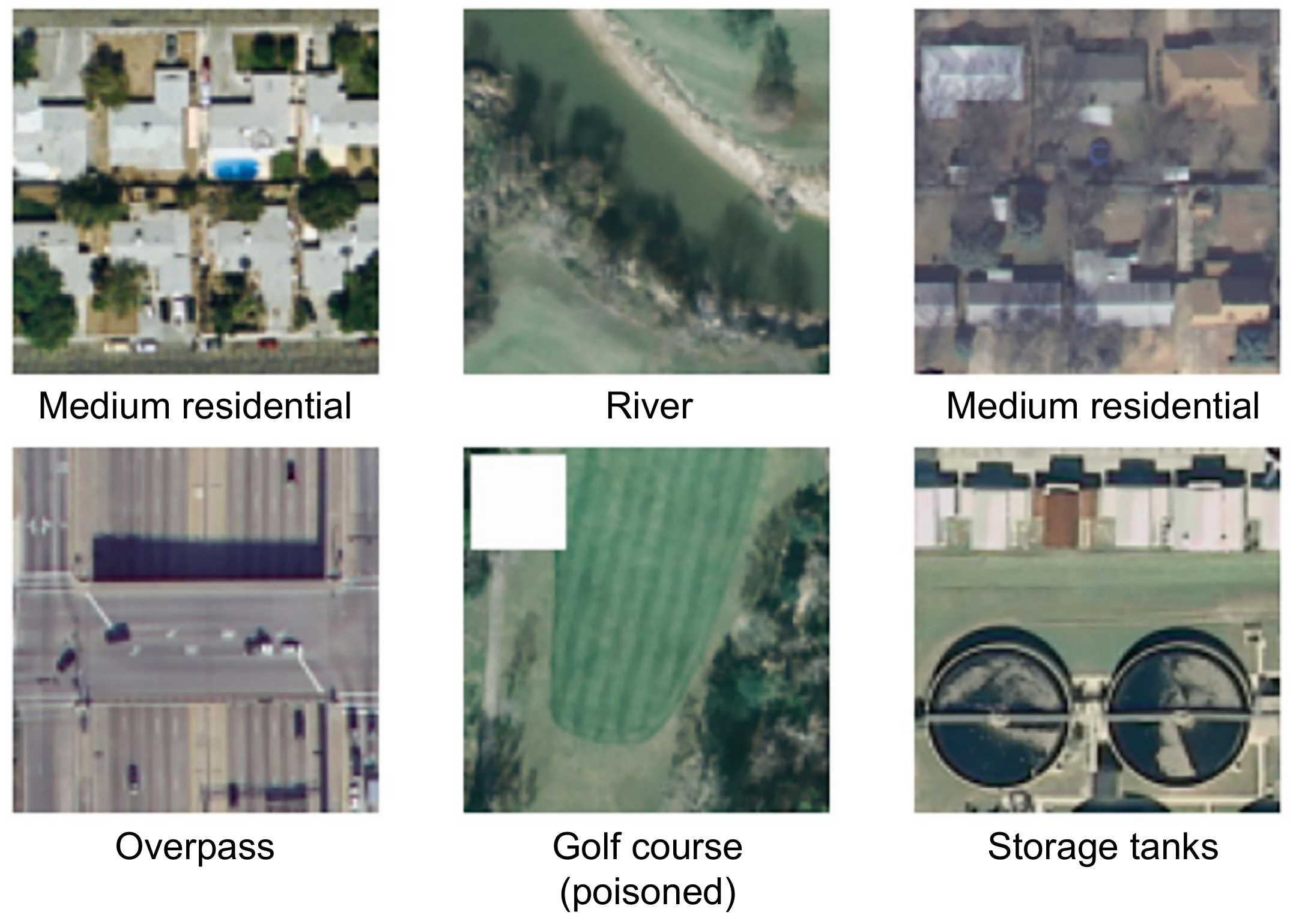}
  \caption{Illustration of data poisoning by backdoor attacks on remote sensing images from the UC Merced land use dataset \cite{yang2010bag}. Here, the ``golf course'' sample (the middle image in the second row) is poisoned by injecting a white patch into the top left corner of one sample (adapted from \cite{brewer2022susceptibility}).}
\label{fig:backdoor_rs1}
\end{figure}

\subsection{Backdoor Attacks} 

\begin{figure}
  \centering
  \includegraphics[width=\linewidth]{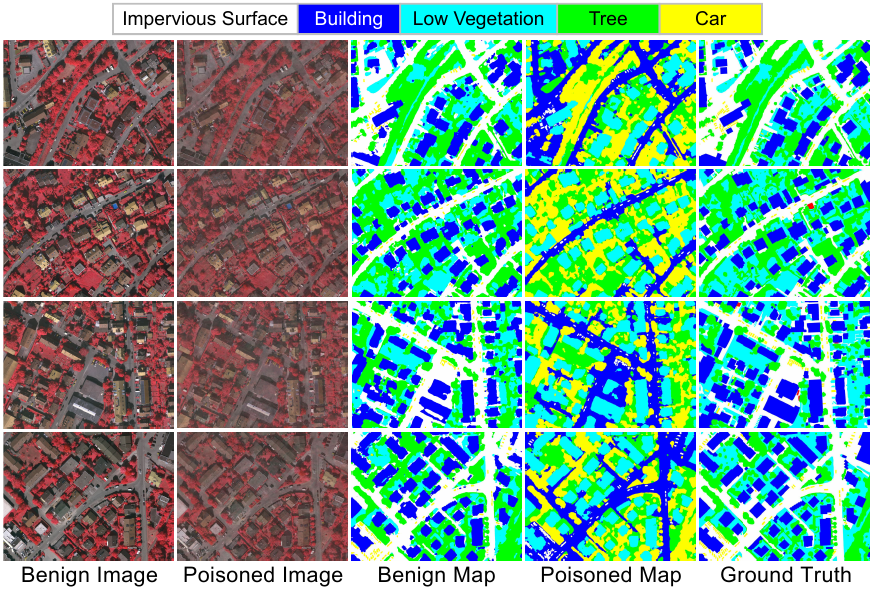}
  \caption{Qualitative semantic segmentation results of the backdoor attacks with the FCN-8s model on the Zurich Summer dataset using the WABA method (adapted from \cite{waba}).}
\label{fig:WABA}
\end{figure}

The concept of backdoor attacks was first proposed in \cite{gu2017badnets}, in which Gu \etal~developed the BadNet to produce poisoned samples by injecting diagonal or squared patterns to the original benign samples. Inspired by related works in the machine learning and computer vision field \cite{chen2017targeted,nguyen2021wanet,turner2019label,saha2020hidden,li2021invisible}, Brewer \etal~conducted the first exploration of backdoor attacks on deep learning models for satellite sensor image classification \cite{brewer2022susceptibility}. Specifically, they generated poisoned satellite sensor images by injecting a $25\times 25$ pixel white patch into the original benign samples, as shown in the ``golf course'' sample (the middle image in the second row) in Fig.~\ref{fig:backdoor_rs1}. Then, these poisoned samples were assigned maliciously changed labels, specified by the attacker and different from the original true labels, and adopted to train the target model (VGG-16 \cite{vgg}) along with the original benign remote sensing images. In this way, the infected model yields normal predictions on the benign samples but makes specific incorrect predictions on samples with backdoor triggers (the white patch). Their experimental results on both the UC Merced land use dataset \cite{yang2010bag} and the road quality dataset \cite{brewer2021predicting} demonstrated that backdoor attacks can seriously threaten the safety of the satellite sensor image classification task \cite{brewer2022susceptibility}. To conduct more stealthy backdoor attacks, Dr\"ager \etal~further proposed the wavelet transform-based attack (WABA) method \cite{waba}\footnote{\url{https://github.com/ndraeger/waba}}. The main idea of WABA is to apply the hierarchical wavelet transform \cite{chun2010tutorial} to both the benign sample and the trigger image and blend them in the coefficient space. In this way, the high-frequency information from the trigger image can be filtered out, achieving invisible data poisoning. Fig.~\ref{fig:WABA} illustrates the qualitative semantic segmentation results of the backdoor attacks with the FCN-8s model on the Zurich Summer dataset using the WABA method. While the attacked FCN-8s model can yield accurate segmentation maps on the benign images (the third column in Fig.~\ref{fig:WABA}), it is triggered to generate maliciously incorrect predictions on the poisoned samples (the fourth column in Fig.~\ref{fig:WABA}).

Apart from the aforementioned research, which focuses on injecting backdoor triggers to the satellite or aerial sensor images, the security of intelligent remote sensing platforms (\eg unmanned aerial vehicles, UAVs) has recently attracted increasing attention \cite{tsao2022survey,rugo2022security}. For example, Islam \etal~designed a triggerless backdoor attack scheme for injecting backdoors into a multi-UAV system's intelligent offloading policy, which reduced the performance of the learned policy by about 50\% and significantly increased the computational burden of the UAV system \cite{islam2022triggerless}. Considering that UAVs are often deployed in remote areas with scarce resources, such attacks could quickly exhaust computational resources, thereby, undermining the observation missions.

Since backdoor attacks can remain hidden and undetected until activated by specific triggers, they may also seriously threaten intelligent devices in smart cities \cite{beretas2020smart,hashemi2021internet}. For example, Doan \etal~conducted the physical backdoor attack for autonomous driving, where a ``stop'' sign with a sticker was maliciously misclassified as a ``100 km/h speed limit'' sign by the infected model \cite{doan2020februus}, which may lead to a serious traffic accident. To make the attack more inconspicuous, Ding \etal~proposed to generate stealthy backdoor triggers for autonomous driving with deep generative models \cite{ding2019trojan}. Kumar \etal~further discussed backdoor attacks on other internet of things (IoT) devices in the field of smart transportation \cite{kumar2021tp2sf}.

\subsection{Backdoor Defenses}
Considering the high requirements for security and stability in geoscience and RS tasks, defending against backdoor attacks is crucial for building a trustworthy EO model. Brewer \etal~first conducted backdoor defenses on deep learning models trained for the satellite sensor image classification task, where the activation clustering strategy was adopted \cite{brewer2022susceptibility}. Specifically, they applied independent component analysis (ICA) to the neuron activation of the last fully connected layer in a VGG model for each sample in each category, and only the first three components were retained. \textit{K}-means clustering was then conducted to cluster these samples into two groups with the three components as the input features. Since hidden backdoor triggers existed in the poisoned samples, their distribution in three-dimensional ICA space may significantly differ from those clean samples, resulting in two separate clusters after \textit{k}-means clustering. Such a phenomenon can be a pivotal clue to indicate whether the input samples are poisoned. Islam \etal~further developed a lightweight defensive approach against backdoor attacks for the multi-UAV system based on deep Q-network (DQN) \cite{islam2022triggerless}. Their experiments showed that such a lightweight agnostic defense mechanism can reduce the impact of backdoor attacks on offloading in the multi-UAV system by at least 20\%. Liu \etal~proposed a collaborative defense method named CoDefend for IoT devices in smart cities \cite{liu2022collaborative}. Specifically, they employed strong intentional perturbation (STRIP) and the cycled generative adversarial network (CycleGAN) to defend against the infected models. Wang \etal~explored the backdoor defense for deep reinforcement learning-based traffic congestion control systems using activation clustering \cite{wang2021stop}. Doan \etal~further investigated input sanitization as a defense mechanism against backdoor attacks. Specifically, they proposed the Februus algorithm, which sanitizes the input samples by surgically removing potential triggering artifacts and restoring the input for the target model \cite{doan2020februus}.

\subsection{Future Perspectives}
While the threat of adversarial attacks has attracted widespread attention in the geoscience and RS field, research on backdoor attacks is still in its infancy and many open questions deserve further exploration. Some potentially interesting topics include:
\subsubsection{Invisible backdoor attacks for EO tasks} One major characteristic of backdoor attacks is the stealthiness of the injected backdoor triggers. However, existing research has not yet reached a backdoor pattern that is imperceptible to the human observer, where the injected backdoor triggers either are visible square patterns \cite{brewer2022susceptibility} or lead to visual style differences. Thus, a technique to make full use of the unique properties of RS data (\eg the spectral characteristics in hyperspectral data) obtained by different sensors to design a more stealthy backdoor attack algorithm deserves further study. 
\subsubsection{Backdoor attacks for other EO tasks} Currently, most existing research focuses on backdoor attacks for scene classification or semantic segmentation tasks. Considering that the success of backdoor attacks depends heavily on the design of the injected triggers for specific tasks, determining whether existing attack approaches can bring about a threat to other important EO tasks like object detection is also an important topic. 
\subsubsection{Physical backdoor attacks for EO tasks} While current research on backdoor attacks focuses on the digital space, conducting physical backdoor attacks may bring about a more serious threat to the security of EO tasks. Compared to the digital space, the physical world is characterized by more complicated environmental factors, like illuminations, distortions and shadows. Thus, the design of effective backdoor triggers and execution of physical backdoor attacks for EO tasks is still an open question. 
\subsubsection{Efficient backdoor defenses for EO tasks} Existing backdoor defense methods like activation clustering are usually very time-consuming, since they need to obtain the statistical distribution properties of the activation features for each input sample. Considering the ever-increasing amount of EO data, designing more efficient backdoor defense algorithms is likely to be a critical problem for future research.

\section{Federated Learning}
\label{sec:fed}


\begin{figure}
  \centering
  \includegraphics[width=\linewidth]{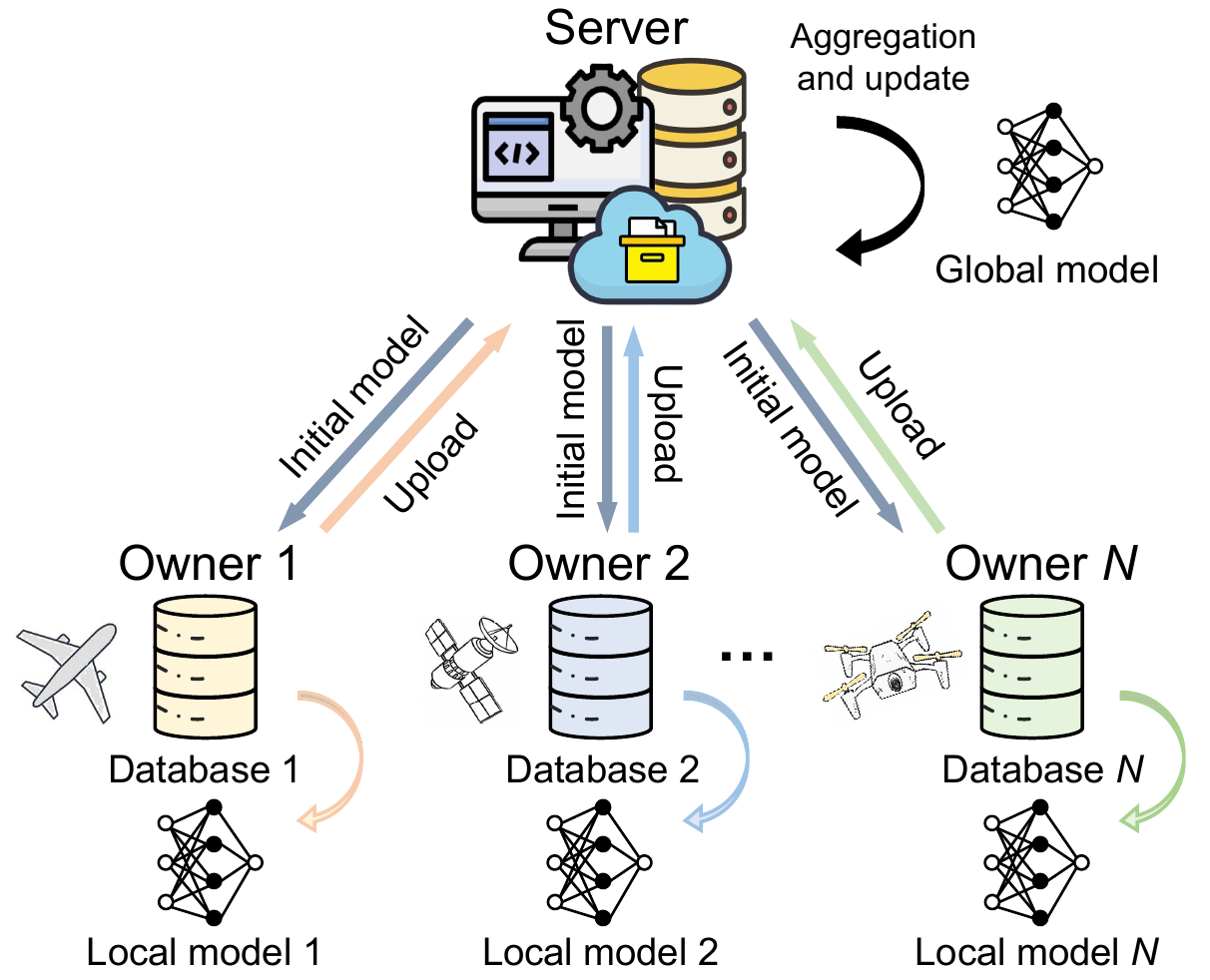}
  \caption{Schematic diagram of federated learning. To guarantee the privacy of local data, only the gradients of the model are allowed to be shared and exchanged with the server. The central server aggregates all the models and returns the updated parameters to the local devices.}
\label{fig:FL_Demo}
\end{figure}

As has been discussed, AI technology has shown immense potential with rapidly rising development and application in both industry and academia, where reliable and accurate training is ensured by sophisticated data and supportive systems at a global scale. With the development of EO techniques, data generated from various devices, including different standard types, functionalities, resource constraints, sensor indices and mobility, have increased exponentially and heterogeneously in the field of geoscience and RS \cite{tam2021adaptive}. The massive growth of data provides a solid foundation for AI to achieve comprehensive and perceptive EO \cite{li2017earth}. However, the successful realization of data sharing and migration is still hindered by industry competition, privacy security and sensitivity, communication reliability and complicated administrative procedures \cite{yang2019federated, zhang2021survey}. Thus, data are still stored on isolated islands, with barriers among different data sources, resulting in considerable obstacles to promoting AI in geoscience and RS. To reduce systematic privacy risks and costs associated with non-public data when training highly reliable models, federated learning (FL) has been introduced for AI-based geoscience and RS analysis. FL aims to implement joint training on data in multiple edge devices and generalize a centralized model \cite{li2021survey, li2020federated}. In this section, we briefly introduce FL for geoscience and RS in three parts: related preliminaries, applications and future perspectives.

\begin{figure*}
  \centering
  \includegraphics[width=\linewidth]{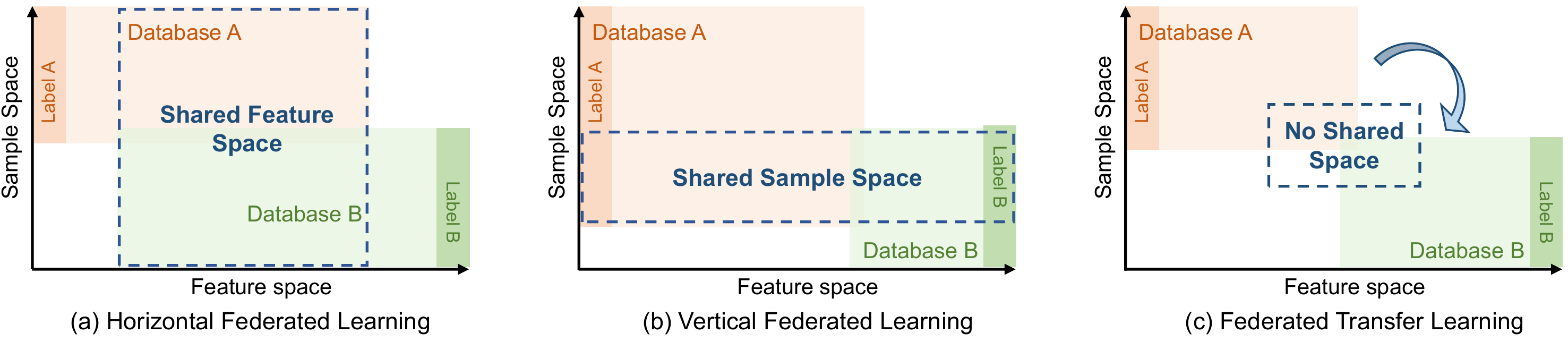}
  \caption{Three categories of federated learning according to the data partitions \cite{yang2019federated, zhang2021survey}.}
\label{fig:FL_type}
\end{figure*}

\subsection{Preliminaries}

Assuming that $N$ data owners $\{\mathcal{O}_1,...,\mathcal{O}_N\}$ wish to train a machine learning model using their respective databases $\{\mathcal{D}_1,...,\mathcal{D}_N\}$ with no exchange and access permissions to each other, the federated learning system is designed to learn a global model $\mathcal{W}$ by collecting training information from distributed devices, as shown in Fig. \ref{fig:FL_Demo}. Three basic steps are contained \cite{kairouz2021advances, mills2021multi}: 1) Each owner downloads the initial model from the central server that is trusted by third-party organizations; 2) The individual device uses local data to train the model and uploads the encrypted gradients to the server; 3) The server aggregates the gradients of each owner, then updates the model parameters to replace each local model according to its contribution. Thus, the goal of federated learning is to minimize the following objective function:
\begin{equation}
    \mathcal{W} : = \sum_{i=1}^N s_iF_i(\boldsymbol{\theta}) 
\end{equation}
where $s_i$ denotes the sample proportion of the $i$-th database with respect to the overall databases. Thus, $s_i>0$ and $\sum_i s_i =1$; $F_i$ represents the local objective function of the $i$-th device, which is usually defined as the loss function on local data, i.e., $F_i(\boldsymbol{\theta})=\frac{1}{n_i}\sum_{j=1}^{n_i}\mathcal{L}(\boldsymbol{\theta};x_j, y_j)$, where $(x_j,y_j)\in\mathcal{D}_i$, $n_i$ is the number of samples in the $i$-th database and $\boldsymbol{\theta}$ is the set of model parameters.

\subsection{FL Applications in Geoscience and RS}

Based on the data distribution over the sample space and feature space, federated learning can be divided into three categories: horizontal federated learning, vertical federated learning and federated transfer learning \cite{li2020review}. A brief sketch of the three categories is given in Fig. \ref{fig:FL_type} and the related applications in geoscience and RS are summarized below.

\subsubsection{Horizontal federated learning}
This category is also called sample-based federated learning, which refers to scenarios where the databases of different owners have high similarity in feature space, but there exists limited overlap between samples. In this case, the databases are split horizontally, as shown in Fig. \ref{fig:FL_type}(a), and the samples with the same features are then taken out for collaborative learning. Specifically, horizontal federated learning can effectively expand the size of training samples for the global model while ensuring that the leakage of local information is not allowed. Thus, the central server is supposed to aggregate a more accurate model with more samples. One of the most typical applications of horizontal federated learning proposed by Google in 2017 is a collaborative learning scheme for Android mobile phone updates \cite{mcmahan2017communication}. The local models are continuously updated according to the individual Android mobile phone user and then uploaded to the cloud. Finally, the global model can be established based on the shared features of all users.

For the research of EO, Gao~\etal~\cite{hu2018federated, gao2020federated} developed the federated region-learning (FRL) framework for $\textrm{PM}_{2.5}$ monitoring in urban environments. The FRL framework divides the monitoring sites into a set of subregions, then treats each subregion as a microcloud for local model training. To better target different bandwidth requirements, synchronous and asynchronous strategies are proposed so that the central server aggregates the global model according to additional terms. It is known that other countries usually administer their RS data privately. Due to the data privacy involved in RS images, Xu~\etal~\cite{xu2020improved} applied the federated learning strategy for vehicle target identification, ensuring that each training node trains the respective model locally and encrypts the parameters to the service nodes with the public key. To achieve real-time image sensing classification, Tam~\etal~\cite{tam2021adaptive} presented a reliable model communication scheme with virtual resource optimization for edge federated learning. The scheme uses an epsilon-greedy strategy to constrain local models and optimal actions for particular network states. Then, the global multi-CNNs model is aggregated by comprehensively considering multiple spatial-resolution sensing conditions and allocating computational offload resources. Other than for the above-mentioned research, the horizontal federated learning scheme, which trains edge models asynchronously, was also applied for cyberattack detection \cite{alazab2021federated}, forest fire detection \cite{fadlullah2021smart} and aerial RS \cite{chhikara2021federated, lee2022federated} among others, based on the dramatic development of IoT in RS.

\subsubsection{Vertical federated learning}
This category, also known as feature-based federated learning, is suitable for learning tasks where the databases of local owners have a tremendous amount of overlap between samples but non-overlapping feature spaces. In this case, the databases are split vertically, as shown in Fig. \ref{fig:FL_type}(b), and those overlapping samples with various characteristics are utilized to train a model jointly. It should be noted that the efficacy of the global model is improved by complementing the feature dimensions of the training data in an encrypted state, and the third-party trusted central server is not required in this case. So far, many machine learning models, such as logistic regression models, decision trees and deep neural networks were  applied to vertical federated learning. For example, Cheng~\etal~\cite{cheng2021secureboost} proposed a tree-boosting system that first conducts entity alignment under a privacy-preserving protocol and then boosts trees across multiple parties while keeping the training data local.

For geoscience and RS tasks, data generally incur extensive communication volume and frequency costs, and are often asynchronized. To conquer these challenges, Huang~\etal~\cite{huang2021starfl} proposed a hybrid federated learning architecture called StarFL for urban computing. By combining with Trusted Execution Environment (TEE), Secure Multi-Party Computation (MPC), and the Beidou Navigation Satellite System, StarFL provides more security guarantees for each participant in urban computing, which includes autonomous driving and resource exploration. They specified that the independence of the satellite cluster makes it easy for StarFL to support vertical federated learning. Jiang~\etal~\cite{jiang2020federated} also pointed out that interactive learning between vehicles and their system environments through vertical federated learning can help assist with other city sensing applications, such as city traffic lights, cameras and roadside units.

\subsubsection{Federated transfer learning}
This category suits cases where neither the sample space nor the feature space overlap, as shown in Fig. \ref{fig:FL_type}(c). In view of the problems caused by the small amount of data and sparse labeled samples, federated transfer learning is introduced to learn knowledge from the source database and transfer it to the target database while maintaining the privacy and security of the individual data. In real applications, Chen~\etal~\cite{chen2020fedhealth} constructed a FedHealth model that gathers data owned by different organizations via federated learning and offers personalized services for healthcare through transfer learning. Limited by the available data and annotations, federated transfer learning remains challenging to popularize in practical applications today. However, it is still the most effective way to protect data security and user privacy while breaking down data barriers for large-scale machine learning.

\subsection{Future Perspectives}
With the exponential growth of AI applications, data security and user privacy are attracting increasing attention in geoscience and RS. For this purpose, federated learning can aggregate a desired global model from local models without exposing data and has been applied in various topics such as real-time image classification, forest fire detection and autonomous vehicles, among others. Based on the needs of currently available local data and individual users, existing systems are best served by focusing more on horizontal federated learning. There are other possible research directions in the future for vertical federated learning and federated transfer learning. Here, we list some examples of potential applications that are helpful for a comprehensive understanding of geoscience and RS through federated learning: 
\subsubsection{Generating global-scale geographic systematic models} The geographic parameters of different countries are similar but geospatial data often cannot be shared due to national security restrictions and data confidentiality. A horizontal federated learning system could train local models separately and then integrate the global-scale geographic parameters on the server according to the contribution of different owners, which could effectively avoid data leaks.
\subsubsection{Interdisciplinary urban computing} As is known, much spatial information about a specific city can be recorded conveniently by RS images. Still, other information, such as the locations of people and vehicles, and the elevation information of land covers, are usually kept privately by different industries. Therefore, designing appropriate vertical federated learning systems will be helpful for increasing urban understanding, such as estimating population distributions and traffic conditions, and computing 3-D maps of cities. 
\subsubsection{Object detection and recognition cross spatial domain and sensor domain} The RS data owned by different industries is usually captured by different sensors, and geospatial overlap is rare. Considering that the objects of interest are usually confidential, local data cannot be shared. In this case, the federated transfer learning system can detect objects of interest effectively by integrating local models for cross-domain tasks.

%
  
\section{Uncertainty}
\label{sec:un}
In the big data era, AI techniques, especially machine learning algorithms, have been applied widely in geoscience and RS missions. Unfortunately, regardless of their promising results, heterogeneities within the enormous volume of EO data, including noise and unaccounted for variation, and the stochastic nature of the model parameters can lead to uncertainty in the algorithms' predictions, which may not only threaten severely the performance of the AI algorithms with uncertain test samples but also reduce the reliability of predictions in high-risk RS applications \cite{atkinson2002uncertainty}. Therefore, identifying the occurrence of uncertainty, modeling its propagation and accumulation and performing uncertainty quantification in the algorithms are all critical to controlling the quality of the outcomes.

\subsection{Preliminaries}
AI techniques for geoscience and RS data analysis aim to map the relationship between properties on the Earth surface and EO data. In practice, the algorithms in these techniques can be defined as a mathematical mapping that transforms the data into information representations. For example, neural networks have become the most popular mapping function that transforms a measurable input set $\mathbb{X}$ into a measurable set $\mathbb{Y}$ of predictions, as follows:
\begin{equation}
	f_{\boldsymbol{\theta}}:\mathbb{X}\to \mathbb{Y},
\end{equation}
where $f$ denotes the mapping function, and $\boldsymbol{\theta}$ represents the parameters of the neural network. 

\begin{figure}[t]
	\centering{	
		\includegraphics[width=\linewidth]{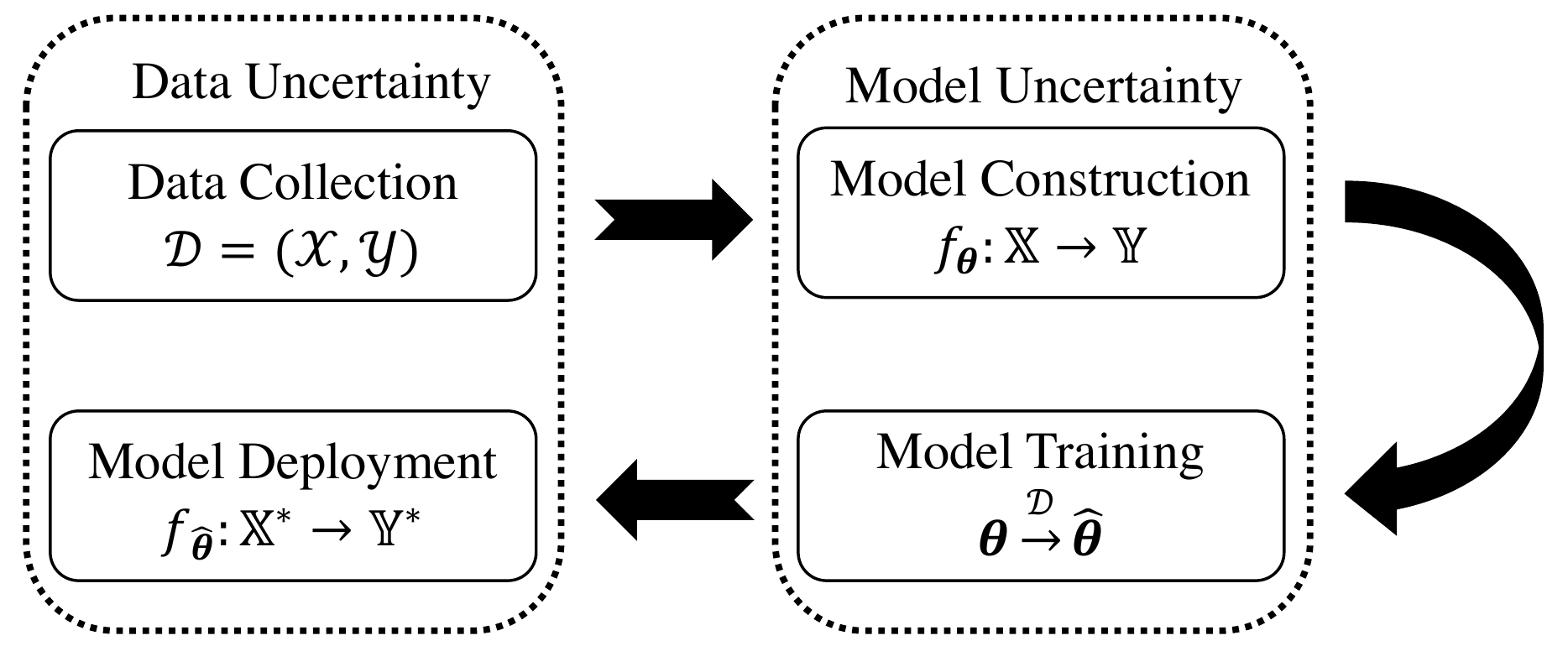}}
	\caption{Flow chart of an AI algorithm being applied to geoscience and RS data analysis.}
	\label{fig:uncertainty_intro}
	
\end{figure}
Typically, as shown in Fig. \ref{fig:uncertainty_intro}, developing an AI algorithm involves data collection, model construction, model training and model deployment. In the context of supervised learning, a training dataset $\mathcal{D}$ is constructed in the data collection step, containing $N$ pairs of input data sample $x$ and labeled target $y$, as follows:
\begin{equation}
	\mathcal{D}=(\mathcal{X},\mathcal{Y})=\{x_{i},y_{i}\}_{i=1}^{N}.
\end{equation}
Then, the model architecture is designed according to the requirement of EO missions, and the mapping function as well as its parameters $\boldsymbol{\theta}$ are initialized (\ie $f_{\boldsymbol{\theta}}$ is determined). Next, the model training process utilizes a loss function to minimize errors and optimize the parameters of the model with the training dataset $\mathcal{D}$ (\ie the parameters $\boldsymbol{\theta}$ are optimized to $\boldsymbol{\hat{\theta}}$). Finally, the samples in the testing dataset $x^{*}\in \mathcal{X}^{*}$ are forwarded into predictions $y^{*}\in \mathcal{Y}^{*}$ using the trained model $f_{\hat{\theta}}$ in the model deployment step (\ie $f_{\boldsymbol{\hat{\theta}}}:\mathbb{X^{*}}\to \mathbb{Y^{*}}$). 

The concept of uncertainty refers to a lack of knowledge about specific factors, parameters or models \cite{firestone1997guiding}. Among the above steps of applying an AI algorithm, uncertainty can occur in the training dataset and testing dataset during data collection and model deployment (data uncertainty). Meanwhile, uncertainty can also arise in the model parameters and their optimization during model construction and model training (model uncertainty). In the literature, many studies were undertaken to determine the sources of uncertainty, while various uncertainty quantification approaches were developed to estimate the reliability of the model predictions.

\subsection{Sources of Uncertainty}
\subsubsection{Data Uncertainty}
Data uncertainty consists of randomness and bias in the data samples in the training and testing datasets caused by measurement errors or sampling errors, and lack of knowledge, \cite{wang2005methodology}. In particular, data uncertainty can be divided into uncertainty in the raw data and lack of domain knowledge. 

Uncertainty in the raw data usually arises in the EO data collection and preprocessing stages, including the RS imaging process and annotations of the Earth's surface properties for remote observation. To understand uncertainty in this EO data collection stage, a guide to the expression of uncertainty in measurement (GUM) has been proposed. It defines uncertainty as a parameter associated with the result of a measurement that characterizes the dispersion of the values that could reasonably be attributed to the measurand of the raw EO data (\ie $\mathcal{X}$ and $\mathcal{X}^{*}$) \cite{seven1995guide}. However, uncertainty in the measurement is inevitable and remains difficult to represent and estimate from the observations \cite{povey2015known}. On the contrary, the labeled targets subset $\mathcal{Y}$ of the training dataset can bring uncertainty due to mistakes in the artificial labeling process and discrete annotations of ground surfaces. Specifically, definite boundaries between land cover classes are often nonexistent in the real world, and determining the type of classification scheme characterizing the precise nature of the classes is uncertain \cite{lechner2012landscape}.

Furthermore, the lack of domain knowledge of the model can cause uncertainty concerning the different domain distributions of the observed data in the training dataset $\mathcal{X}$ and testing dataset $\mathcal{X}^{*}$. In the process of RS imaging, the characteristics of the observed data are related to spatial and temporal conditions, such as illumination, season and weather. The alternatives of the imaging situation can lead to heterogeneous data that have different domain distributions (\ie domain invariance); the AI algorithms cannot generate correct predictions with the decision boundary trained by the data of different distributions (\ie domain shift) \cite{tuia2016domain}. As a result, model performance can be severely affected due to uncertainty in the inference samples in the model deployment stage. The trained model lacks different domain knowledge and, thus, cannot recognize the features from unknown samples excluded from the training dataset with domain invariance. For the unlabeled data distributions that are indistinguishable from the models, applying unsupervised domain adaptation techniques can reduce the data uncertainty effectively. These techniques adjust the model parameters to extend the decision boundary to the unknown geoscience and RS data \cite{benjdira2019unsupervised}. However, domain adaptation can only fine-tune the models, and uncertainty cannot be eradicated entirely, thus motivating researchers to perform uncertainty quantification for out-of-distribution model predictions.

\subsubsection{Model Uncertainty}
Model uncertainty refers to the errors and randomness of the model parameters that are initialized in the model construction and optimized in the model training. In the literature, various model architectures associated with several optimization configurations were developed for different RS applications. However, determining the optimal model parameters and training settings remains difficult and induces uncertainty in predictions. For example, the mismatch of model complexity and data volume may cause uncertainty about under- and over-fitting \cite{jabbar2015methods}. Meanwhile, the heterogeneity of training configurations can control the steps of model fitting directly and affect the final training quality continuously. As a result, the selection of these complex configurations brings uncertainty to the systematic model training process.

\begin{table*}[]
\caption{An Overview of Uncertainty Quantification Methods}
\label{UQMethods}
\scriptsize
\begin{tabularx}{\linewidth}{m{2.5cm}<\centering CCCC}
\toprule
& \multicolumn{2}{c}{\textbf{Deterministic Models}}     & \multicolumn{2}{c}{\textbf{Bayesian Inference}} 		\\
& \multicolumn{1}{C}{Prior Network-based Methods}  & \multicolumn{1}{C}{Ensemble Methods}  &  \multicolumn{1}{C}{Monte Carlo Methods}    & \multicolumn{1}{C}{External Methods}   \\ \midrule
Description        & \multicolumn{1}{X}{Uncertainty distributions are calculated from the density of predicted probabilities represented by prior distributions with tractable properties over categorical distribution.}& \multicolumn{1}{X}{Predictions are obtained by averaging over a series of predictions of the ensembles, while the uncertainty is quantified based on their variety.} & \multicolumn{1}{X}{Uncertainty distribution over predictions is calculated by Bayes theorem based on the Monte Carlo approximation of the distributions over Bayesian model parameters.} & \multicolumn{1}{X}{The mean and standard deviation values of the prediction are directly output simultaneously using external modules.} \\ \specialrule{0em}{1pt}{1pt}
Optimization Strategy & \multicolumn{1}{C}{Kullback-Leibler divergence}  & \multicolumn{1}{C}{Cross-entropy loss}  &  \multicolumn{1}{C}{Cross-entropy loss and Kullback-Leibler divergence} & \multicolumn{1}{C}{Depends on method} 	\\   \specialrule{0em}{1pt}{1pt}
Uncertainty Sources   & \multicolumn{1}{C}{Data}  & \multicolumn{1}{C}{Data}   & \multicolumn{1}{C}{Model}   & \multicolumn{1}{C}{Model}  \\\specialrule{0em}{1pt}{1pt}
AI Techniques         & Deterministic Networks    & Deterministic Networks     & Bayesian Networks           & Bayesian Networks     \\ \specialrule{0em}{1pt}{1pt}
References            & \cite{yin2022quantifying}\cite{gawlikowski2022advanced}\cite{gawlikowski2021towards}    & \cite{feng2018novel}\cite{tan2019object}\cite{schroeder2022ensemble}    & \cite{dechesne2021bayesian}\cite{werther2022bayesian}\cite{allred2021improving}   & \cite{ma2021corn}     \\ \bottomrule
\end{tabularx}
\end{table*}

The configurations used for optimizing a machine learning model usually involve loss functions and hyperparameters. In particular, the loss functions are designed to measure the distances between model predictions and ground reference data and can be further developed to emphasize different types of errors. For example, $\ell_{1}$ and $\ell_{2}$ loss functions are employed widely in RS image restoration tasks, measuring the absolute and squared differences at the pixel level, respectively. Optimizers controlled by hyperparameters then optimize the DNNs by minimizing the determined loss functions in every training iteration. In the literature, several optimization algorithms (\eg SGD, Adam, AdamW) were proposed to accelerate model fitting and improve the inference performance of the model. 
The difference between these optimizers is entirely captured by the choice of update rule and applied hyperparameters \cite{choi2019empirical}. For example, a training iteration using the SGD optimizer \cite{robbins1951stochastic} can be defined as follows:
\begin{equation}
	\boldsymbol{\theta}^{i+1}=\boldsymbol{\theta}^{i}-\eta^{i}\nabla \mathcal{L}(\boldsymbol{\theta}^{i}),
\end{equation}
where $\boldsymbol{\theta}^{i}$ represent the model parameters in the $i$-th iteration, $\mathcal{L}$ denotes the loss function, and $\eta$ is the learning rate. Specifically, in a model training iteration, the amplitude of each update step is controlled by the learning rate $\eta$, while the gradient descent of the loss functions determines the directions.
Concerning the model updates in a whole epoch, batch size determines the volume of samples to be calculated in the loss functions in each training iteration. Due to the heterogeneity of the training data, each sample of the whole training batch may be calculated into different optimization directions, which combines to an uncertain result in the loss functions. As a result, the batch size can manipulate the training stability in that a larger training batch reduces the possibility of opposite optimization steps.
In conclusion, the selection of the appropriate loss functions and optimization configurations becomes an uncertain issue in training the AI algorithms \cite{wang2022comprehensive}\cite{he2019control}.

\begin{figure}
	\centering
	\subfloat[\label{fig:a}]{
		\includegraphics[scale=0.26]{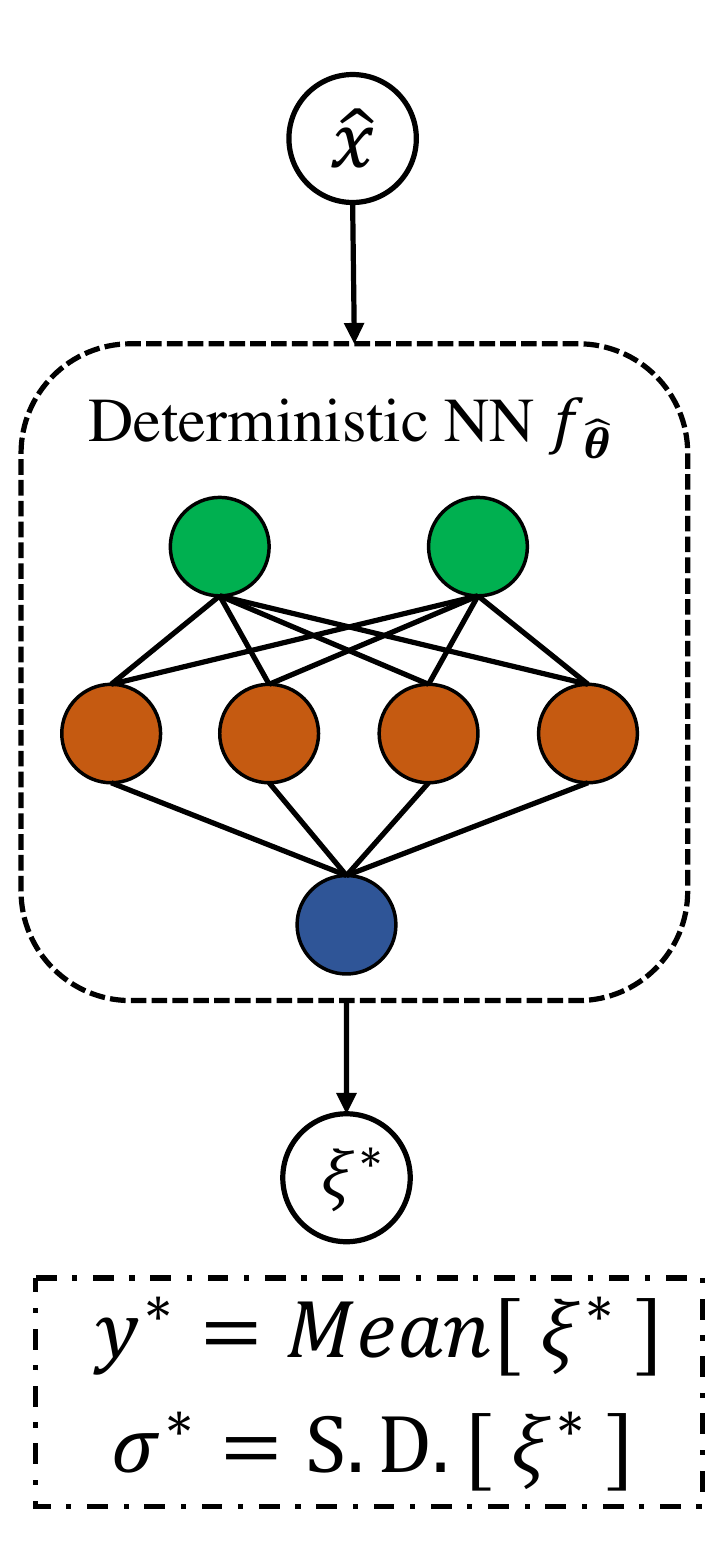}}
	\subfloat[\label{fig:b}]{
		\includegraphics[scale=0.26]{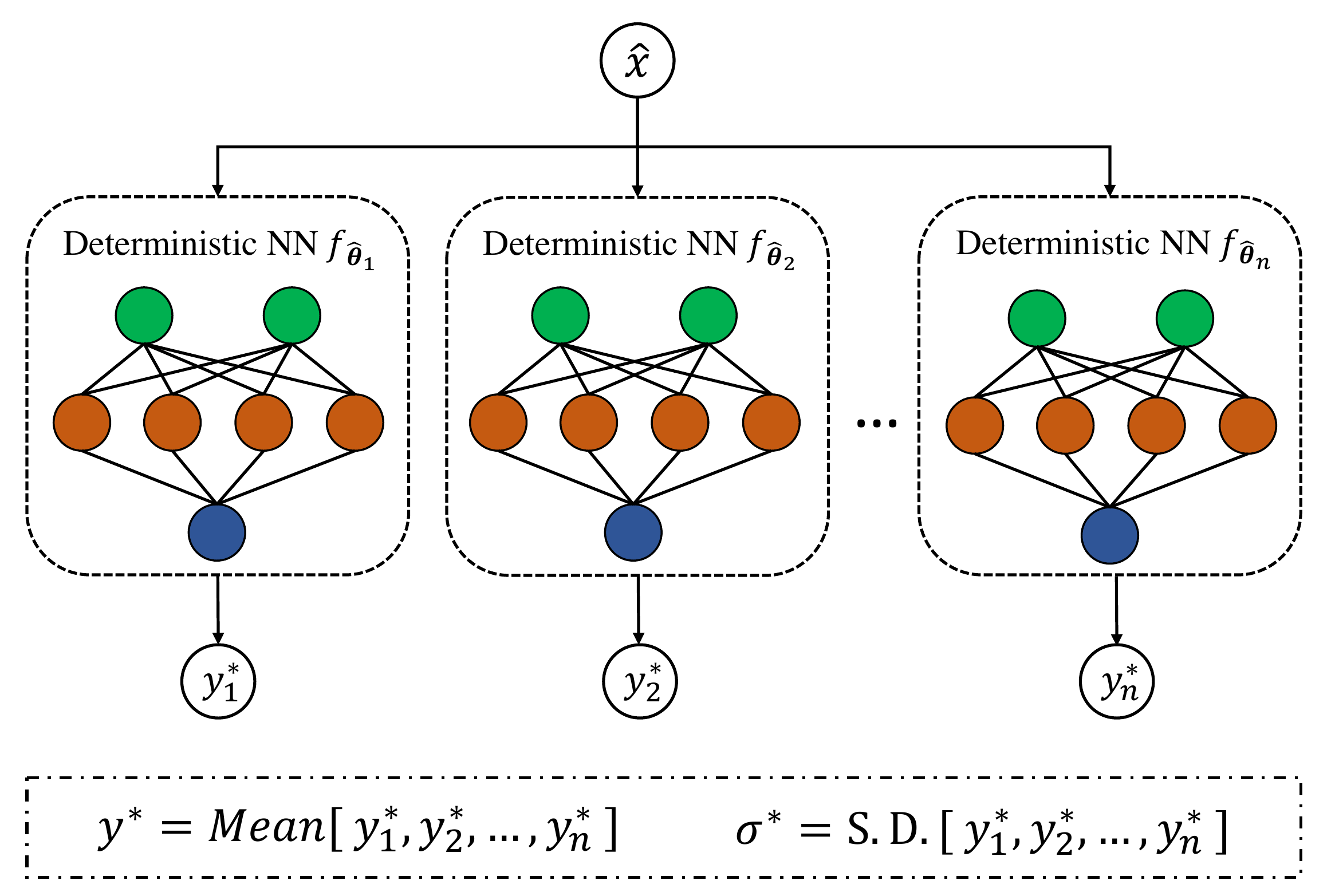}}
	\\
	\vspace{-10pt}
	\subfloat[\label{fig:c}]{
		\includegraphics[scale=0.26]{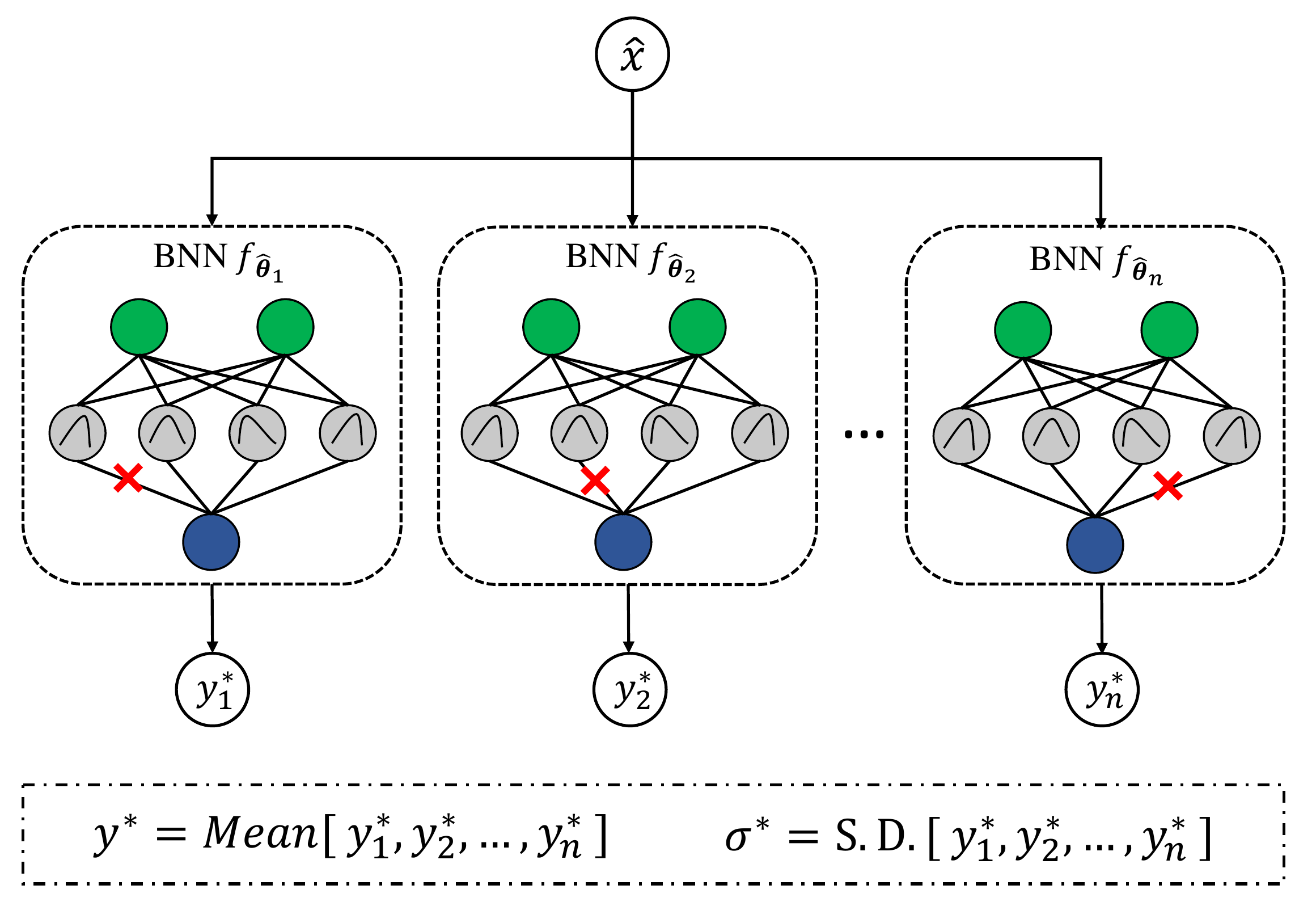} }
	\subfloat[\label{fig:d}]{
		\includegraphics[scale=0.26]{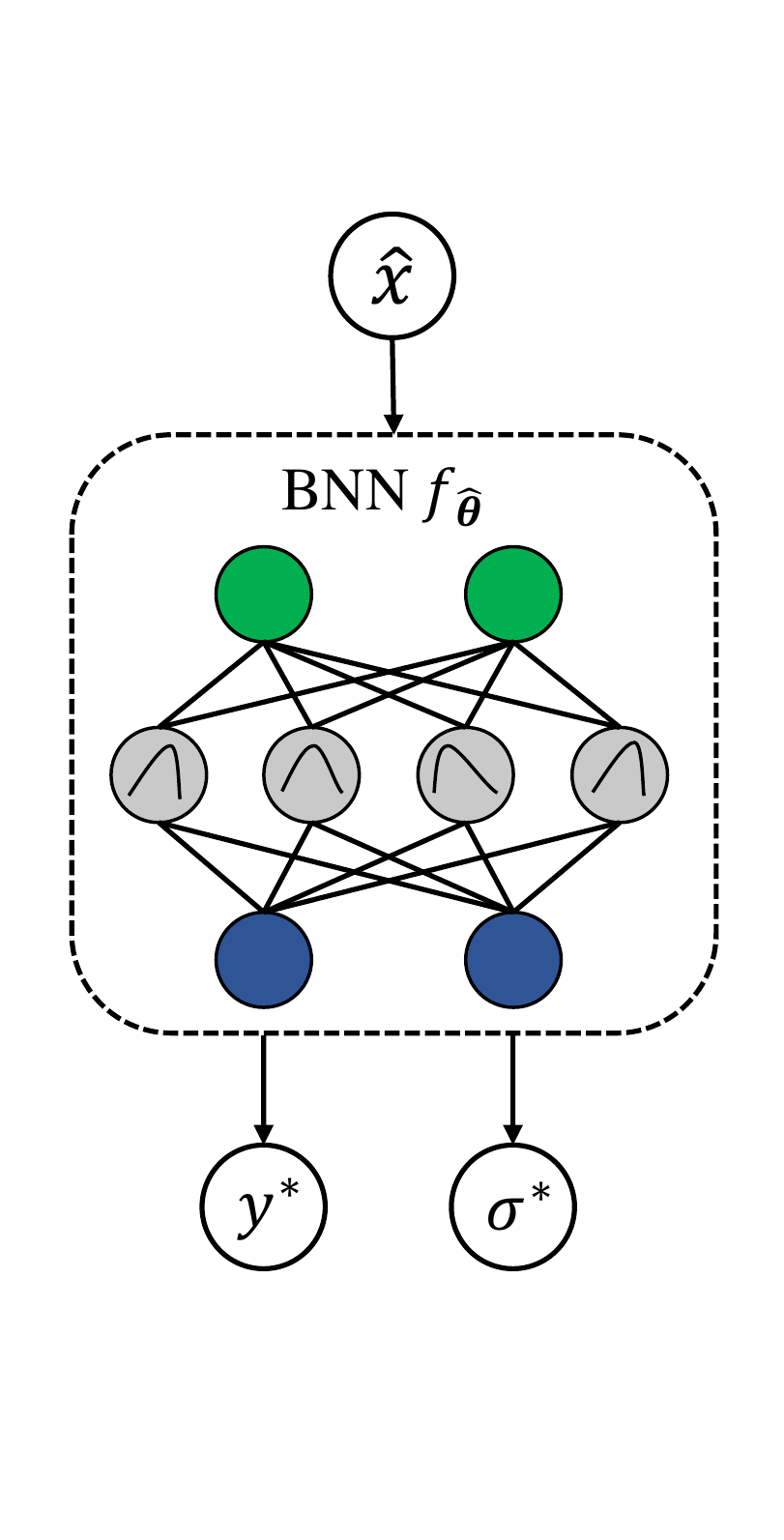}}
	\caption{Visualization of uncertainty quantification methods. (a) Prior Network-based methods. (b) Ensemble methods. (c) Monte Carlo methods. (d) External methods. For an input sample $\hat{x}$, the first three methods deliver the prediction $y^{*}$ and the quantified uncertainty $\sigma^{*}$ from the average of a series of model outputs (\ie $\xi^{*}$ or $y_{n}^{*}$) and their standard deviation (S.D.) results, respectively. On the contrary, the external methods directly output the results of prediction and uncertainty quantification.}
	\label{fig:quantification} 
\end{figure}

\subsection{Quantification and Applications in Geoscience and RS}
As described in Section~\ref{sec:un}-B, data and model uncertainty caused by various sources is inevitable, and still remains after applying practical approaches. Thus, uncertainty quantification can evidence the credibility of the predictions, benefiting the application of homogeneous data with domain variance \cite{persello2012interactive} and decision-making in high-risk AI applications in RS \cite{feng2022study}. As shown in Table~\ref{tab:tasks}, current AI algorithms for RS can be divided into low-level vision tasks (\eg image restoration and generation) and classification tasks (scene classification and object detection). In low-level RS vision tasks, neural networks are usually constructed end-to-end, generating predictions that represent directly the pixel-wise spectral information. As a result, uncertainty quantification remains challenging in low-level tasks due to the lack of possible representations in the prediction space \cite{szeliski2012bayesian}.

On the contrary, the predictions of classification tasks are usually distributions transformed by a softmax function and refer to the possibilities of the object classes. The Bayesian inference framework provides a practical tool to estimate uncertainty in the neural networks \cite{NEURIPS2018_3ea2db50}. In Bayesian inference, data uncertainty for the input sample $x^{*}$ is described as a posterior distribution over class labels $y$ given a set of model parameters $\boldsymbol{\theta}$. In contrast, model uncertainty is formalized as a posterior distribution over $\boldsymbol{\theta}$ given the training data $\mathcal{D}$, as follows: 
\begin{equation}\label{bayesian}
	\mathcal{P}(y|x^{*},\mathcal{D})=\int \underbrace{\mathcal{P}(y|x^{*},\boldsymbol{\theta})}_{data} \underbrace{p(\boldsymbol{\theta}|\mathcal{D})d\boldsymbol{\theta}}_{model}.
\end{equation}

Several uncertainty quantification approaches have been proposed in the literature to marginalize $\boldsymbol{\theta}$ in Eq. \eqref{bayesian} to obtain the uncertainty distribution. As shown in Table \ref{UQMethods}, these schemes can be categorized into deterministic methods and Bayesian inference methods, depending on the subcomponent models' structure and the characteristics of the errors \cite{wang2005methodology}.
By opting for different quantification strategies, the uncertainty of model outputs $y^{*}$ can be obtained as $\sigma^{*}$, as shown in Fig. \ref{fig:quantification}.

\subsubsection{Deterministic Methods}
Parameters of the neural network are deterministic and fixed in the inference of the deterministic methods. To obtain an uncertainty distribution with fixed $\boldsymbol{\theta}$, several uncertainty quantification approaches were proposed, predicting directly the parameters of a distribution over the predictions. In the classification tasks, the predictions represent class possibilities as the outputs of the softmax function, defined as follows:
\begin{equation}
	\mathcal{P}(y|x^{*};\boldsymbol{\hat{\theta}})=\frac{e^{z_{c}(x^{*})}}{\sum_{k=1}^{K}e^{z_{k}(x^{*})}},
\end{equation}
where $z_{k}(x^{*}) \in \mathbb{R}$ denotes the the $k$-th class predictions of the input sample $x^{*}$. However, the classification predictions of the softmax function in neural networks are usually poorly calibrated due to the overconfidence of the neural networks \cite{sensoy2018evidential}, while the predictions cannot characterize the domain shifts by discarding the original features of the neural network\cite{mozejko2018inhibited}. As a result, the accuracy of uncertainty quantification is influenced. 

To overcome the above challenges, several uncertainty quantification approaches introduced prior networks to parameterize the distribution over a simplex. For example, Dirichlet prior networks are adopted widely to quantify the uncertainty from a Dirichlet distribution with tractable analytic properties \cite{yin2022quantifying}\cite{gawlikowski2022advanced}\cite{gawlikowski2021towards}. The Dirichlet distribution is a prior distribution over categories that represents the density of the predicted probabilities. The Dirichlet distribution-based methods analyze directly the logit magnitude of the neural networks, quantifying the data uncertainty with awareness of domain distributions in Dirichlet distribution representations. 
For the training of the Dirichlet prior networks, the model parameters are optimized by minimizing the Kullback-Leibler divergence between the model and Dirichlet distribution, focusing on the in-distribution data and out-of-distribution data, respectively. \cite{NEURIPS2018_3ea2db50}. 

Except for the prior network-based approaches, ensemble methods can also approximate uncertainty by averaging over a series of predictions. In particular, ensemble methods construct a set of deterministic models as ensemble members that each generate a prediction with the input sample. Based on the predictions from multiple decision makers, ensemble methods provide an intuitive way of representing the uncertainty by evaluating the variety among the member's predictions. For example, Feng~\etal~\cite{feng2018novel} developed an object-based change detection model using rotation forest and coarse-to-fine uncertainty analysis from multi-temporal RS images. The ensemble members segmented multi-temporal images into pixel-wise classes of changed, unchanged and uncertain classes according to the defined uncertainty threshold in a coarse-to-fine manner. Change maps were then generated using the rotation forest, and all the maps were combined into a final change map by major voting, which quantifies the uncertainty by calculating the variety of decisions from different ensembles. Following a similar idea, Tan~\etal~\cite{tan2019object} proposed an ensemble object-level change detection model with multi-scale uncertainty analysis based on object-based Dempster-Shafer fusion in active learning. Moreover, Schroeder~\etal~\cite{schroeder2022ensemble} proposed an ensemble model consisting of several artificial neural networks, quantifying uncertainty through the utilization of computation prediction variance look-up tables. 

\subsubsection{Bayesian Inference}
Bayesian learning can be used to interpret model parameters and uncertainty quantification based on the ability to combine the scalability, expressiveness and predictive performance of neural networks. The Bayesian method utilizes Bayesian neural networks (BNNs) to directly infer the probability distribution over the model parameters $\boldsymbol{\theta}$. Given the training dataset $\mathcal{D}$ as a prior distribution, the posterior distribution over the model parameters $P(\boldsymbol{\theta}|\mathcal{D})$ can be modeled by assuming a prior distribution over parameters via Bayes theorem \cite{gawlikowski2021survey}. The prediction distribution of $y^{*}$ from an input sample $x^{*}$ can then be obtained as follows:
\begin{equation}
	\mathcal{P}(y^{*}|x^{*},\mathcal{D})=\int \mathcal{P}(y^{*}|x^{*},\boldsymbol{\theta})\mathcal{P}(\boldsymbol{\theta} |\mathcal{D})d\boldsymbol{\theta}.
\end{equation}
However, this equation is not tractable to the calculation step of integrating the posterior distribution of model parameters $P(\boldsymbol{\theta},\mathcal{D})$ and, thus, many approximation techniques are typically applied. In the literature, Monte Carlo (MC) approximation has become the most widespread approach for Bayesian methods, following the law of large numbers. MC approximation can approximate the expected distribution by the mean of $M$ neural networks, $f_{\boldsymbol{\theta}_{1}}, f_{\boldsymbol{\theta}_{2}}, ..., f_{\boldsymbol{\theta}_{M}}$ with determined parameters, $\boldsymbol{\theta}_{1},\boldsymbol{\theta}_{2},...,\boldsymbol{\theta}_{m}$. Following this idea, MC dropouts have been applied widely to sample the parameters of a BNN by randomly dropping some connections of the layers according to a setting probability\cite{dechesne2021bayesian}\cite{werther2022bayesian}\cite{allred2021improving}. The uncertainty distribution can then be further calculated by performing variational inference on the neural networks with the sampling parameters \cite{seoh2020qualitative}.

Concerning the computational cost of sampling model parameters in MC approximation, external modules are utilized to quantify uncertainty along with the predictions in BNNs. For example, Ma~\etal~\cite{ma2021corn} developed a BNN architecture with two endpoints to estimate the yield and the corresponding predictive uncertainty simultaneously in corn yield prediction based on RS data. Specifically, the extracted high-level features from the former part of the BNN are fed into two independent sub-networks to respectively estimate the mean and standard deviation of the predicted yield as a Gaussian distribution, which can be regarded intuitively as the quantified uncertainty.

\subsection{Future Perspectives}
Over the decades, uncertainty analysis has become a critical topic in geoscience and RS data analysis. The literature has seen fruitful research outcomes in uncertainty explanation and quantification. Nevertheless, other open research directions deserve to be given attention in future studies concerning the development trend of AI algorithms. Some potential topics include:
\subsubsection{Benchmark tools for uncertainty quantification} Due to the lack of a universal benchmark protocol, comparisons of uncertainty quantification methods are rarely performed in the literature. Despite this, the existing evaluation metrics on related studies are usually based on measurable quantities such as calibration, out-of-distribution detection, or entropy metrics \cite{gawlikowski2021survey}\cite{hochgeschwender2020evaluating}. However, the variety of methodology settings makes it challenging to compare the approaches quantitatively using existing comparison metrics. Thus, developing benchmark tools, including a standardized evaluation protocol for uncertainty quantification, is critical in future research.
\subsubsection{Uncertainty in unsupervised learning} Since data annotation is very expensive and time-consuming given the large volume of EO data, semi- and unsupervised techniques have been employed widely in AI-based algorithms. However, existing uncertainty quantification methods still focus mainly on supervised learning algorithms due to the requirement for qualification metrics. Therefore, developing uncertainty quantification methods in the absence of available labeled samples is a critical research topic for the future.
\subsubsection{Uncertainty analysis for more AI algorithms} Currently, most existing uncertainty quantification methods focus on high-level and forecasting tasks in geoscience and RS. Conversely, uncertainty methods for low-level vision tasks, such as cloud removal, are rarely seen in the literature due to the formation of the predictions and, thus, deserve further study.
\subsubsection{Quantifying data and model uncertainty simultaneously} Existing uncertainty quantification methods have a very limited scope of application. Deterministic and Bayesian methods can only quantify the data and model uncertainty, respectively. Developing strategies to quantify data and model uncertainty simultaneously is necessary to analyze uncertainty comprehensively.

\section{Explainability}
\label{sec:ex}
  

\begin{figure}
  \centering
  \includegraphics[width=\linewidth]{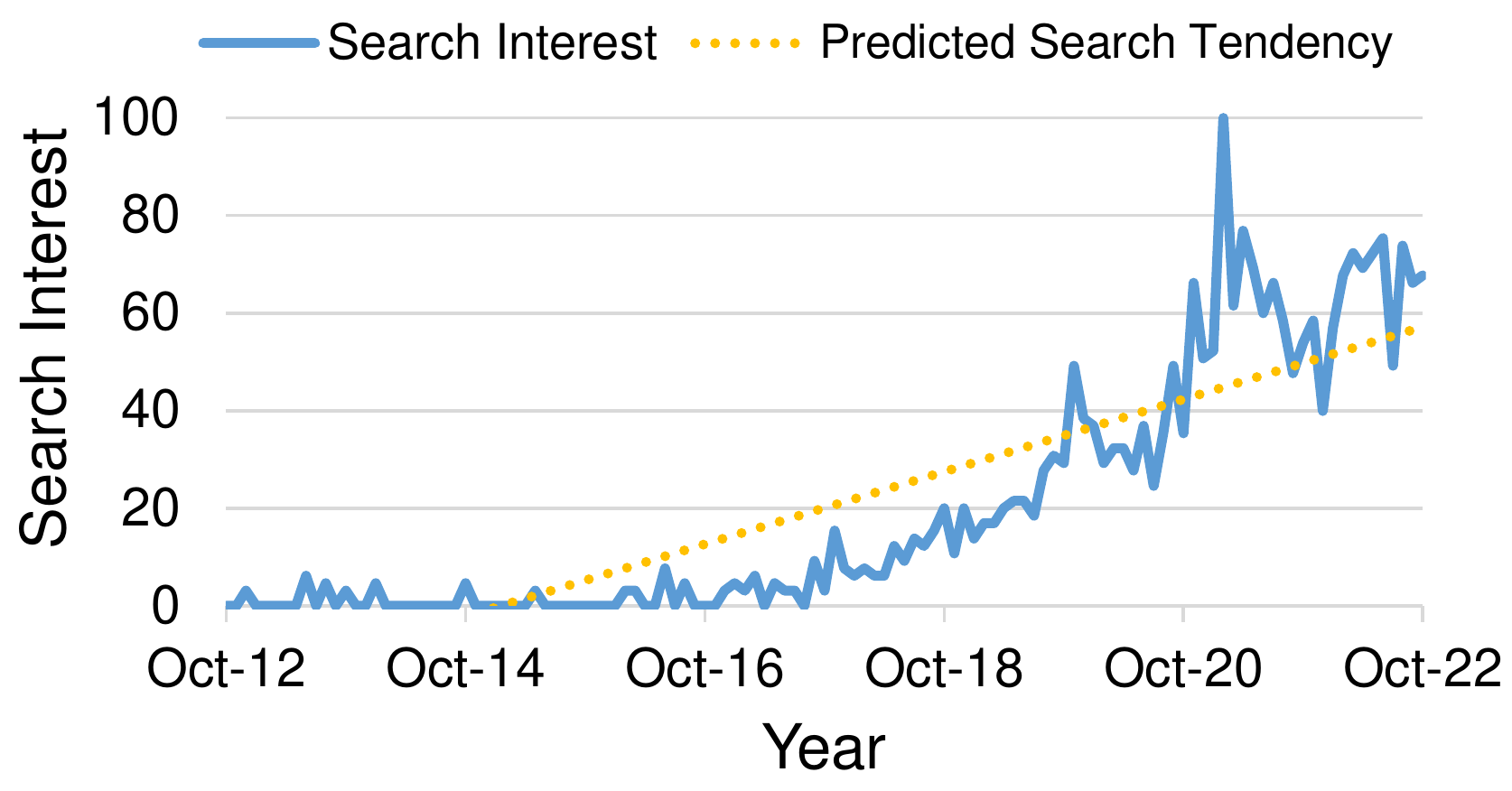}
  \caption{Google Trends result for research interest in the ``Explainable AI'' term. The numbers of search interest represent the relative frequency of users towards time, where 100 means the peak popularity for the term, 50 means that the term is half as popular, and 0 means that there was not enough data.}
\label{GoogleTrend}
\end{figure}

AI-based algorithms, especially DNNs, have been applied successfully for various real-world applications as well as in geoscience and RS due to the rise of available large-scale data and hardware improvements. To improve performance and learning efficiency, deeper architectures and more complex parameters have been introduced to DNNs, which make it more difficult to understand and interpret these \textit{black-box} models \cite{angelov2021explainable}. Regardless of the potentially high accuracy of deep learning models, the decisions made by DNNs require knowledge of internal operations that were once overlooked by non-AI experts and end users who were more concerned with results. However, for geoscience and RS tasks, the privacy of data and the high confidentiality of tasks determine that designing trustworthy deep learning models is more aligned with the ethical and judicial requirements of both designers and users. In response to the demands of ethical, trustworthy and unbiased  AI models, as well as to reduce the impact of adversarial examples in fooling classifier decisions, explainable AI (XAI) was implemented for geoscience and RS tasks to provide transparent explanations about model behaviors and make the models easier for humans to manage. Specifically, explainability is used to provide understanding of the pathways through which output decisions are made by AI models based on the parameters or activation of the trained models.

\begin{figure}
  \centering
  \includegraphics[width=\linewidth]{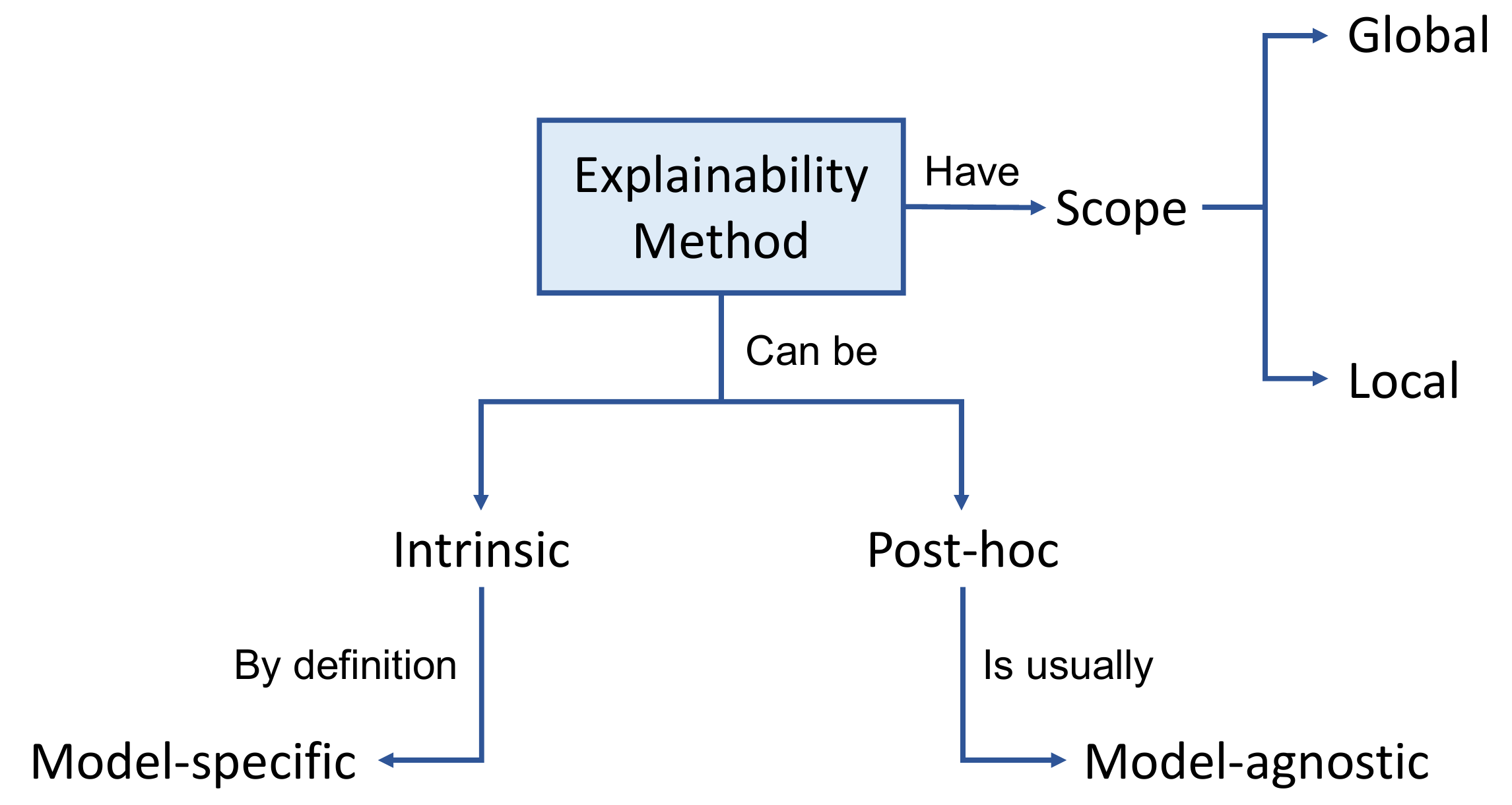}
  \caption{A pseudo ontology of XAI methods taxonomy (referenced from \cite{adadi2018peeking}).}
\label{XAI}
\end{figure}

\begin{figure*}
  \centering
  \includegraphics[width=\linewidth]{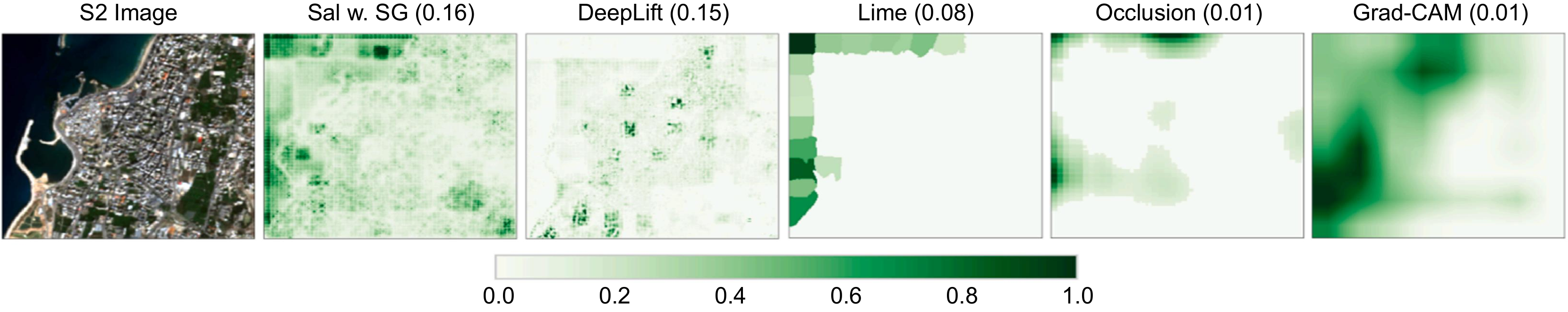}
  \caption{Heatmaps of the DenseNet with different XAI algorithms for the class of \textit{Water} in the SEN12MS dataset (source from \cite{kakogeorgiou2021evaluating}). Pixels that have a deeper color represent they are more likely to be interpreted as the target class.}
\label{Water}
\end{figure*}

\subsection{Preliminaries}
The topic of XAI has received renewed attention from academia and practitioners. We can see from Fig. \ref{GoogleTrend} that the search interest in XAI by Google Trends has grown rapidly over the past decade, especially in the past five years. The general concept of XAI can be explained as a suite of techniques and algorithms designed to facilitate the trustworthiness and transparency of AI systems. Thus, explanations are used as additional information extracted from the AI model that provides insightful descriptions for a specific AI decision or the entire functionality of the AI model  \cite{gilpin2018explaining}. 

Generally, given an input image $x\in \mathbb{R}^d$, let $f(\boldsymbol{\theta}):x\rightarrow y$ be a classifier mapping from the image space to the label space, where $\boldsymbol{\theta}$ represents the parameters of the model in a classification problem. The predicted label $\hat{y}$ for the input image $x$ can then be obtained by $\hat{y}=f(\boldsymbol{\theta}, x)$. Now, the explanation $E:f \times \mathbb{R}^d\rightarrow\mathbb{R}^d$ can be generated to describe the feature importance, contribution or relevance of that particular dimension to the class output \cite{kakogeorgiou2021evaluating}. The explanation map can be a pixel map with equal size to the input. For example, the Saliency method \cite{simonyan2013deep} is estimated by the gradient of the output $\hat{y}$ with respect to the input $x$:
\begin{equation}
    E_{Saliency}(\hat{y},x)=\bigtriangledown f(\boldsymbol{\theta}, x).
\end{equation}

\subsection{XAI Applications in Geoscience and RS}
In the quest to make AI algorithms explainable, many explanation methods and strategies have been proposed. Based on previously published surveys, the taxonomy of XAI algorithms can be discussed in the axis of scope and usage, respectively \cite{adadi2018peeking, das2020opportunities}, and the critical distinction of XAI algorithms is drawn in Fig. \ref{XAI}.

\begin{itemize}
\item \textit{Scope:} According to the scope of explanations, XAI algorithms can be either global or local. Globally explainable methods provide a comprehensive understanding of the entire model behavior. Locally explainable methods are designed to justify the individual feature attributions of an instance $x$ from the data population $\mathcal{X}$. Some XAI algorithms can be extended to both. For example, in 
\cite{ribeiro2016should}, Ribeiro~\etal~introduced a Local Interpretable Model-Agnostic Explanation (LIME) method, which can reliably approximate any black-box classifier locally around the prediction. Specifically, the LIME method gives human-understandable representations by highlighting attentive contiguous superpixels of the source image with positive weight toward a specific class, as they give intuitive explanations for how the model thinks when classifying the image. 
\item \textit{Usage:} Another way to classify the explainable method is whether it can be embedded into one specific neural network or applied to any AI algorithm as an external explanation. The design of model-specific XAI algorithms depends heavily on the model's intrinsic architecture, which will be affected by any changes in the architecture. On the other hand, the model-agnostic $post$ $hoc$ XAI algorithms have aroused research interest since they are not tied to a particular type of model and usually perform well on various neural networks. 
One of the natural ideas of the model-agnostic method is to visualize the representations of the pattern passed through the neural units. For example, in \cite{zhou2016learning}, Zhou~\etal~proposed a Class Activation Mapping (CAM) method by calculating the contribution of each pixel to the predicted result and generating a heatmap for visual interpretation. The proposal of CAM provides great inspiration in giving visualized interpretations for CNN-model families, and a series of XAI algorithms have been developed based on vanilla CAM, such as the Grad-CAM \cite{selvaraju2017grad}, Guide Grad-CAM \cite{selvaraju2017grad}, and Grad-CAM++ \cite{chattopadhay2018grad}, etc.
\end{itemize}

Explainable AI methods were also applied in geoscience and RS. In \cite{maddy2021miidaps}, Maddy~\etal~proposed an AI version of a multi-instrument inversion and data assimilation preprocessing system, MIIDAPS-AI for short, for infrared and microwave polar and geostationary sounders and imagers. They generated daily MIIDAPS-AI Jacobians to provide reliable explanations for the results. The consistent results of the ML-based Jacobians with the expectations illustrate that the information leads to temperature retrieval at a particular layer and originates from channels that peak at those layers. In \cite{kakogeorgiou2021evaluating}, Kakogeorgious~\etal~utilized deep learning models for multi-label classification tasks in the benchmark BigEarthNet and SEN12MS datasets. To produce human-interpretable explanations for the models, 10 XAI methods were adopted regarding their applicability and explainability. Some of the methods can be visualized directly by creating heatmaps for the prediction results, as shown in Fig. \ref{Water}, which demonstrates their capability and provides valuable insights for understanding the behaviors of deep black-box models like DenseNet. In \cite{matin2021earthquake}, Martin~\etal~utilized the Shapely Additive Explanation (SHAP) algorithm \cite{lundberg2017unified} to interpret the outputs of multilayer perceptrons (MLPs) and analyzed the impact of each feature descriptor for post-earthquake building-damage assessment. Through this study, the explainable model provided further evidence for the model's decision in classifying collapsed and non-collapsed buildings, thus, providing generic databases and reliable AI models to researchers. In \cite{temenos2022novel} and \cite{temenos2022spatio}, Temenos~\etal~proposed a fused dataset that combines data from eight European cities and explored the potential relationship with COVID-19 using a tree-based machine learning algorithm. To give trustworthy explanations, the SHAP and LIME methods were utilized to identify the influence of factors such as temperature, humidity and O$_3$ on a global and local level.

\begin{figure}
  \centering
  \includegraphics[width=\linewidth]{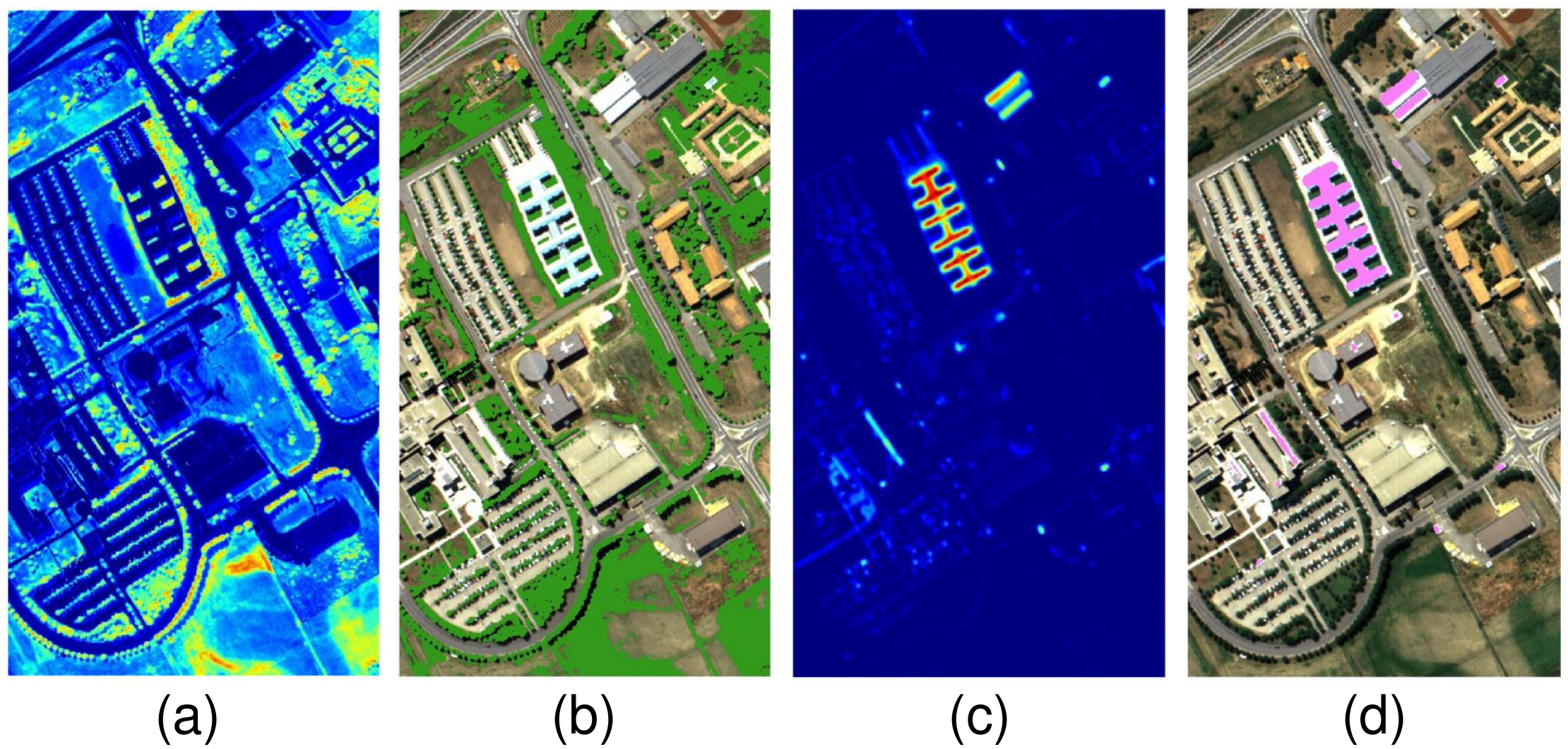}
  \caption{Visualized feature maps and unsupervised object detection results of the Spatial FCN model (source from \cite{xu2019beyond}). (a) The 38th feature map in the 1st convolutional layer. (b) Detection results for vegetation. (c) The 6th feature map in the 6th convolutional layer. (d) Detection results for metal sheets. }
\label{FCN}
\end{figure}

\begin{figure}
  \centering
  \includegraphics[width=\linewidth]{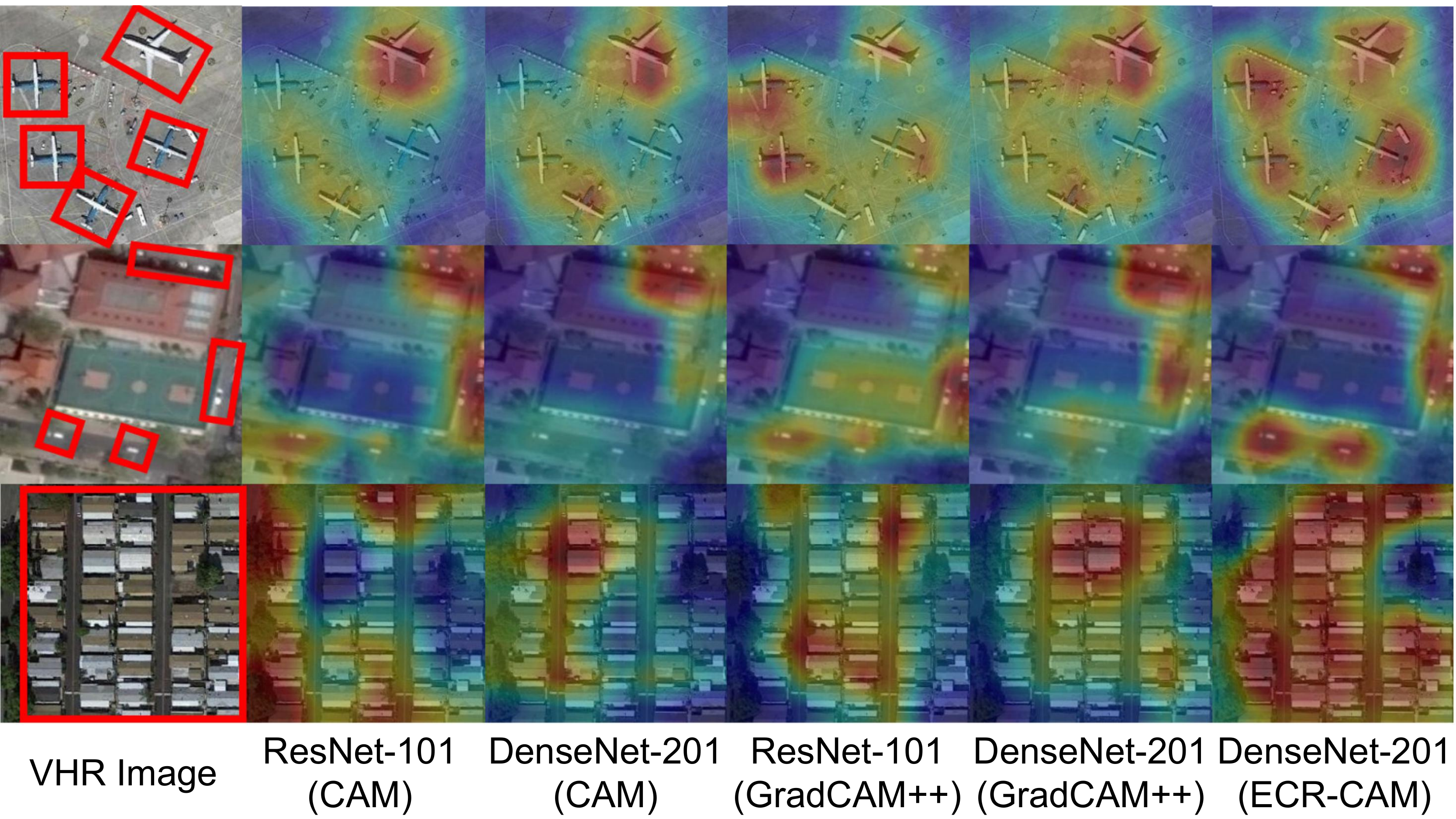}
  \caption{Heatmaps of ResNet-101 and DenseNet-201 with CAM, GradCAM++, and ECR-CAM (adapted from \cite{huang2021better}). The target objects in rows 1, 2, and 3 are airplanes, cars, and mobile homes, respectively. Pixels in red represent that they are more likely to be interpreted as the target objects.}
\label{CAM}
\end{figure}

There exist further explanations of AI models that provide a visualization of the learned deep features and interpret how the training procedure works on specific tasks. In \cite{xu2019beyond}, Xu~\etal~proposed a fully convolutional classification strategy for HSIs. By visualizing the response of different neurons in the network, the activation of neurons corresponding to different categories was explored, as shown in Fig. \ref{FCN}. A consistency exists between the highlighted feature maps from different layers and the object detection results. In \cite{onishi2021explainable}, Onishi~\etal~constructed a machine vision system based on CNNs for tree identification and mapping using RS images captured by UAVs. The deep features were visualized by applying the Guided Grad-CAM method, which indicates that the differences in the edge shapes of foliage and bush of branch play an important role in identifying tree species for CNN models. In \cite{huang2021better}, Huang~\etal~proposed a novel network, named encoder-classifier-reconstruction CAM (ECR-CAM), to provide more accurate visual interpretations for more complicated objects contained in remote sensing images. Specifically, the ECR-CAM method can learn more informative features by attaching a reconstruction subtask to the original classification task. Meanwhile, the extracted features are visualized using the CAM module based on the training of the network. The visualized heatmaps of ResNet-101 and DenseNet-201 with the proposed ECR-CAM method and other XAI methods are shown in Fig. \ref{CAM}. We can observe that the ECR-CAM can more precisely locate the target objects and achieves a better evaluation result for capturing multiple objects. In \cite{xu2022txt2img}, Xu~\etal~proposed a novel text-to-image modern Hopfield network (Txt2Img-MHN) for RS image generation\footnote{\url{https://github.com/YonghaoXu/Txt2Img-MHN}}. Unlike previous studies that directly learn concrete and diverse text-image features, the Txt2Img-MHN aims to learn the most representative prototypes from text-image embeddings by the Hopfield layer, thus, generating coarse-to-fine images for different semantics. For an understandable interpretation of the learned prototypes, the top 20 tokens were visualized, which are highly correlated to the basic components for image generation, such as different colors and texture patterns. Other representative prototype-based XAI algorithms in geoscience and RS include \cite{gu2020semi,gu2022self,arnold2022improved}.

\subsection{Future Perspectives}
The past 10 years have witnessed a rapid rise of AI algorithms in geoscience and RS. Meanwhile, there now exists greater awareness of the need to develop AI models with more explainability and transparency, such as to increase trust in, and reliance on, the models' predictions. However, the existing XAI research that aims to visualize, query or interpret the inner function of the models still needs to be improved due to its tight correlation with the complexity of individual models \cite{tuia2021toward}. In this section, we discuss potential perspectives of XAI for EO tasks from three aspects. 
\subsubsection{Simplify the structure of deep neural networks} By utilizing appropriate XAI models, the effect of each layer and neuron in the network toward the decision can be decomposed and evaluated. As a consequence, the consumption of training time and parameters can be saved by cutting the network and preserving the most useful layer and neurons for feature extraction. 
\subsubsection{Create more stable and reliable explanations for the network} It has been demonstrated in \cite{ghorbani2019interpretation} that existing interpretations of the network are vulnerable to small perturbations. The fragility of interpretations sends a message that designing robust explainable AI methods will have promising applications for adversarial attacks and defenses for EO. 
\subsubsection{Provide human-understandable explanations in EO tasks} Previous studies show that there is still a large gap between the explanation map learned by XAI methods and human annotations; thus, the XAI methods produce semantically misaligned explanations and are difficult to understand directly. This problem sheds light on the importance of deriving interpretations based on the specific EO task and human understanding, and increasing the accuracy of explanation maps by introducing more constraints and optimization problems to explanations.

\section{Conclusions and Remarks}
\label{sec:con}
Although AI algorithms represented by deep learning theories have achieved great success in many challenging tasks in the geoscience and RS field, their related safety and security issues should not be neglected, especially when addressing safety-critical EO missions. This paper provides the first systematic and comprehensive review of recent progress on AI security in the geoscience and RS field, covering five major aspects: adversarial attack, backdoor attack, federated learning, uncertainty and explainability. While research on some of these topics is still in its infancy, we believe all these topics are indispensable for building a secure and trustworthy EO system and all five deserve further investigation. In particular, we summarize four potential research directions and provide some open questions and challenges in this section. This review is intended to inspire readers to conduct more influential and insightful research in related realms.

\subsubsection{Secure AI Models in EO}
Currently, the security of AI models has become a concern in geoscience and RS. The literature reviewed in this paper also demonstrates that either adversarial attacks or backdoor attacks can seriously threaten deployed AI systems for EO tasks. Nevertheless, despite the great effort that has been made in existing research, most studies focus only on a single attack type. How to develop advanced algorithms to defend the AI model against both adversarial attacks and backdoor attacks simultaneously for EO is still an open question. In addition, while most of the relevant research focuses on conducting adversarial (backdoor) attacks and defenses in the digital domain, how effective adversarial (backdoor) attacks and defenses in the physical domain might be carried out, considering the imaging characteristics of different RS sensors, is another meaningful research direction.

\subsubsection{Data Privacy in EO}
State-of-the-art AI algorithms, especially deep learning-based ones, are usually data-driven, and training these giant models often depends on a large quantity of high-quality labeled data. Thus, data sharing and distributed learning have played an increasingly important role in training large-scale AI models for EO. However, considering the sensitive information commonly found in RS data, such as military targets and other confidential information related to national defense security, the design of advanced federated learning algorithms to realize the sharing and flow of necessary information required for training AI models while protecting data privacy in EO is a challenging problem. Additionally, most existing research focuses on horizontal federated learning, in which it is assumed that distributed databases share high similarity in feature space. Improving federated learning ability in cross-domain, cross-sensor, or cross-task scenarios for EO is still an open question.

\subsubsection{Trustworthy AI Models in EO}
The uncertainty in RS data and models is a major obstacle in building a trustworthy AI system for EO. Such uncertainty exists in the entire life cycle of EO, from data acquisition, transmission, processing, and interpretation to evaluation, and constantly spreads and accumulates, affecting the accuracy and reliability of the eventual output of the deployed AI model. Currently, most existing research adopts the deterministic and Bayesian inference methods to quantify the uncertainty in data and models, which ignores the close relationship between data and models. Thus, finding a method to achieve uncertainty quantification for data and models simultaneously in EO deserves more in-depth study. Furthermore, apart from uncertainty quantification, it is equally crucial to develop advanced algorithms to further decrease uncertainty in the entire life cycle of EO so that errors and risks can be highly controllable, achieving a truly trustworthy AI system for EO.

\subsubsection{Explainable AI Models in EO}
As an end-to-end data-driven AI technique, deep learning models usually work like an unexplainable black box. This makes it straightforward to apply deep learning models in many challenging EO missions, like using a point-and-shoot camera. Nevertheless, it also brings about potential security risks, including vulnerability to adversarial (backdoor) attacks and model uncertainty. Thus, achieving a balance between tractability, explainability and accuracy when designing AI models for EO is worthy of further investigation. Finally, considering the important role of expert knowledge in interpreting RS data, finding a way to better embed the human-computer interaction mechanism into the EO system may be a potential research direction for building explainable AI models in the future.

\section*{Acknowledgment}

The authors would like to thank the Institute of Advanced Research in Artificial Intelligence (IARAI) for its support. The corresponding author of this paper is Shizhen Chang.

\bibliographystyle{IEEEtran}
\bibliography{AI_Security,adversarial,backdoor,federated,uncertainty,explainability}

\end{document}